\documentclass[11pt]{article}
\usepackage{amssymb}

\usepackage{cuted}
\usepackage{capt-of} 
\usepackage{float}
\usepackage[preprint]{acl}
\usepackage{booktabs}
\usepackage{tabularx}
\usepackage{makecell}
\usepackage{array}

\usepackage{times}
\usepackage{latexsym}
\usepackage{multirow}
\usepackage{makecell}
\usepackage[T1]{fontenc}

\usepackage[utf8]{inputenc}
\usepackage{amsmath}

\usepackage{microtype}
\usepackage{placeins}
\usepackage{inconsolata}

\usepackage{graphicx}

\title{Learning to Retrieve User History and Generate User Profiles for Personalized Persuasiveness Prediction}

\author{
  \textbf{Sejun Park} \quad
  \textbf{Yoonah Park} \quad
  \textbf{Jongwon Lim} \quad
  \textbf{Yohan Jo\thanks{Corresponding author.}} \\
  Graduate School of Data Science, Seoul National University \\
  \texttt{\{aprimelonge, wisdomsword21, elijah0430, yohan.jo\}@snu.ac.kr}
}

\begin{document}
\maketitle
\begin{abstract}
Estimating the persuasiveness of messages is critical in various applications, from recommender systems to safety assessment of LLMs. While it is imperative to consider the target persuadee's characteristics, such as their values, experiences, and reasoning styles, there is currently no established systematic framework to optimize leveraging a persuadee's past activities (e.g., conversations) to the benefit of a persuasiveness prediction model. To address this problem, we propose a context-aware user profiling framework with two trainable components: a \emph{query generator} that generates optimal queries to retrieve persuasion-relevant records from a user's history, and a \emph{profiler} that summarizes these records into a profile to effectively inform the persuasiveness prediction model.
Our evaluation on the ChangeMyView Reddit dataset shows consistent improvements over existing methods across multiple predictor models, raising F1 from 33\% to 47\% on \texttt{Llama-3.3-70B-Instruct}.
Further analysis shows that effective user profiles are context-dependent and predictor-specific, rather than relying on static attributes or surface-level similarity.
Together, these results highlight the importance of task-oriented, context-dependent user profiling for personalized persuasiveness prediction. Our code and data are available at \url{https://github.com/holi-lab/ReCAP}.
\end{abstract}

\section{Introduction}

Large language models (LLMs) are increasingly used in decision-support applications that aim to influence human behavior or beliefs, such as health coaching, tutoring, and targeted marketing \citep{salvi2024conversational,hackenburg2025levers}. In these settings, an LLM may generate or evaluate multiple candidate messages (e.g., campaign messages for marketing companies) to assist a human decision maker, requiring the system to determine which message is most likely to persuade a target user. We refer to this problem as \emph{persuasiveness prediction}, defined as predicting a user’s belief or attitude change in response to a given message \citep{Perloff2021}. The main challenge in persuasiveness prediction stems from the fact that persuasion is inherently personalized: the same argument may be compelling for one user but ineffective for another, depending on factors such as beliefs, values, experiences, and reasoning style \citep{lukin2017argument,durmus-cardie-2018-exploring,al2020exploiting}. 
As a result, accurate persuasiveness prediction requires inferring how each individual user is likely to interpret and respond to a message.
In practice, such inference must rely on signals from a user’s past interaction history, as explicit user attributes are often unavailable.
This motivates methods that infer user characteristics from historical interactions to enable personalized persuasiveness prediction.

\begin{figure}[t]
    \centering
    \includegraphics[width=\columnwidth]{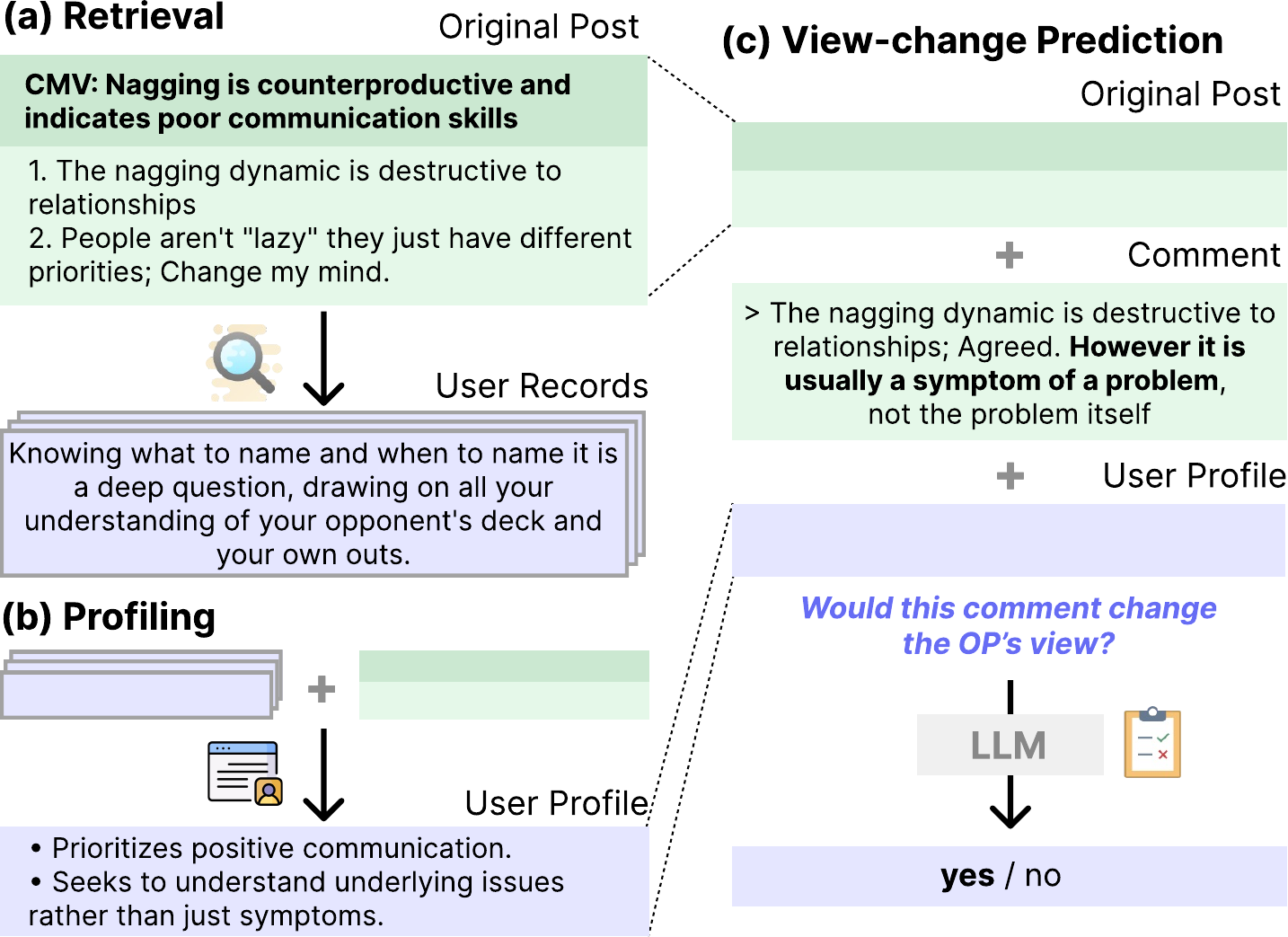}
    \caption{Overview of the view-change prediction task with context-aware user profiling on CMV.
Given an original post, the system (a) retrieves relevant user records, (b) constructs a textual user profile, and (c) predicts whether a comment will change the user’s view.}
    \label{fig:intro}
\end{figure}

We formulate this problem as a \emph{personalized view-change prediction} task using data from the \texttt{ChangeMyView} (CMV) Reddit forum.
In this setting, each instance consists of (i) an original post expressing a user’s view on a specific topic, (ii) a comment responding to the post with the intent to change that view, and (iii) the user’s historical \emph{records}, consisting of past Reddit posts and comments.
The goal is to predict whether the comment will successfully change the user’s stated view by leveraging user information inferred from the user’s history (Figure~\ref{fig:intro}c; \citealp{tan-etal-2016-winning}).
To make accurate predictions, a system requires to construct user representations from their historical records that capture what matters for the current persuasion context (Figure~\ref{fig:intro}a,b; \citealp{li-etal-2016-persona,xu-etal-2025-personalized}).
However, existing approaches typically rely on heuristic retrieval methods (e.g., selecting recent records or random sampling) to select relevant historical records and generic profiling techniques (e.g., extracting demographic traits) to summarize user characteristics from those records \citep{hackenburg2025levers, al2020exploiting, salvi2024conversational}. We argue that these approaches are insufficient because persuasion is inherently context-dependent: which aspects of a user’s history are informative depends on the topic and stance of the original post as well as the argument in the candidate message \citep{tan-etal-2016-winning,ji-etal-2018-incorporating}.

To address this limitation, we propose a framework with two learnable modules for generating a user profile in textual form (Figure~\ref{fig:intro}) : (i) a \emph{query generator} that produces retrieval queries to identify persuasion-relevant records from a user’s history, and (ii) a \emph{profiler} that summarizes the retrieved records into a textual user profile, conditioned on the original post. This profile, together with the original post, is then used by a predictor model to determine whether the candidate message would change the user's view.
Examples of the final user profiles are presented in Appendix~\ref{sec:profile-examples}.

We train the components through the joint optimization of a query generator and a profiler in the following three steps. First, we train the \emph{profiler} to leverage user records to generate effective textual profiles for view-change prediction. Second, building on this profiler, we score each user record based on the effectiveness of its resulting profiles for view-change prediction. Third, we train the \emph{query generator} to retrieve high-scoring user records from the user history pool.

Our evaluation on the CMV dataset shows consistent gains over prior approaches, demonstrating the effectiveness of our framework for personalized persuasiveness prediction, with further gains on OpinionQA and PRISM confirming that the benefits extend beyond CMV.
Further analyses show that persuasion-relevant user characteristics vary across posts and predictor models, highlighting the need for predictor-specific, context-aware user profiles rather than generic or static attributes. Our retrieval-based design also exhibits computational advantages; compared to many baselines that use a summary of the entire user history, our method keeps per-instance inference cost 6--13× lower. Beyond view-change prediction, these findings suggest that effective personalization requires learning both what to retrieve from a user's history and how to summarize it for the given context.

Our contributions are threefold:
\begin{enumerate}
\item \textbf{Annotation-free learnable pipeline.} We train retrieval and profiling via view-change prediction as a utility-based signal, without ground-truth annotations.
\item \textbf{Persuasion-aware query generation.} We retrieve user records by context-relevant aspects rather than semantic similarity.
\item \textbf{Context- and predictor-dependent profiling.} We show effective profiling is dynamic and predictor-specific, not static or predictor-agnostic.
\end{enumerate}

\section{Related Work}

\paragraph{Early Work on Personalized Persuasion}
Since \citet{tan-etal-2016-winning} established the view-change prediction task on the CMV dataset, subsequent work has enriched modeling by incorporating richer linguistic features, such as interaction dynamics and discourse relations \citep{ji-etal-2018-incorporating,hidey-mckeown-2018-persuasive}.
More and more research demonstrated the importance of personalization by leveraging persuadee characteristics in persuasion outcome prediction, including ideology, demographic, and personality  traits \citep{lukin2017argument, durmus-cardie-2018-exploring,durmus-cardie-2019-corpus,durmus-cardie-2019-factors, al2020exploiting}.
However, they largely rely on pre-defined, explicit user attributes.
Leveraging recent advances in LLMs, we infer richer persuasion-relevant user information from the users’ past writings.

\paragraph{User Profiling for LLM Personalization}

Early work has formulated LLM personalization as making models behave \emph{like} a specific user given their historical writings \citep{salemi2024lamp, mysore2024pearl}.
Building on this, several studies have explored retrieval- and profiling-based approaches \citep{richardson2023integrating, li2024learning, salemi2024optimization, zhang2024guided}.
These methods focus on linguistic style and topical relevance, remaining limited in capturing user \emph{values} \citep{qin-etal-2025-similarity}, which are crucial for personalized persuasion.
Studies on personalized dialogue agents construct user profiles via summarization to generate user-aligned responses \citep{zhong2024memorybank, wang2025recursively}, but remain limited in dynamically adapting profiles to the current interaction context.

Another line of work fine-tunes the predictor on users’ historical data to encode user-specific information \citep{zhang2025prime}.
While effective, it requires retraining as new user data arrives, limiting scalability in practice.
In contrast, we keep the predictor fixed and focus on constructing context-aware user profiles, and thus do not directly compare against such approaches.

\section{User Profiling Framework}
\label{sec:method}

\begin{figure*}[t]
    \centering
    \includegraphics[width=\textwidth]{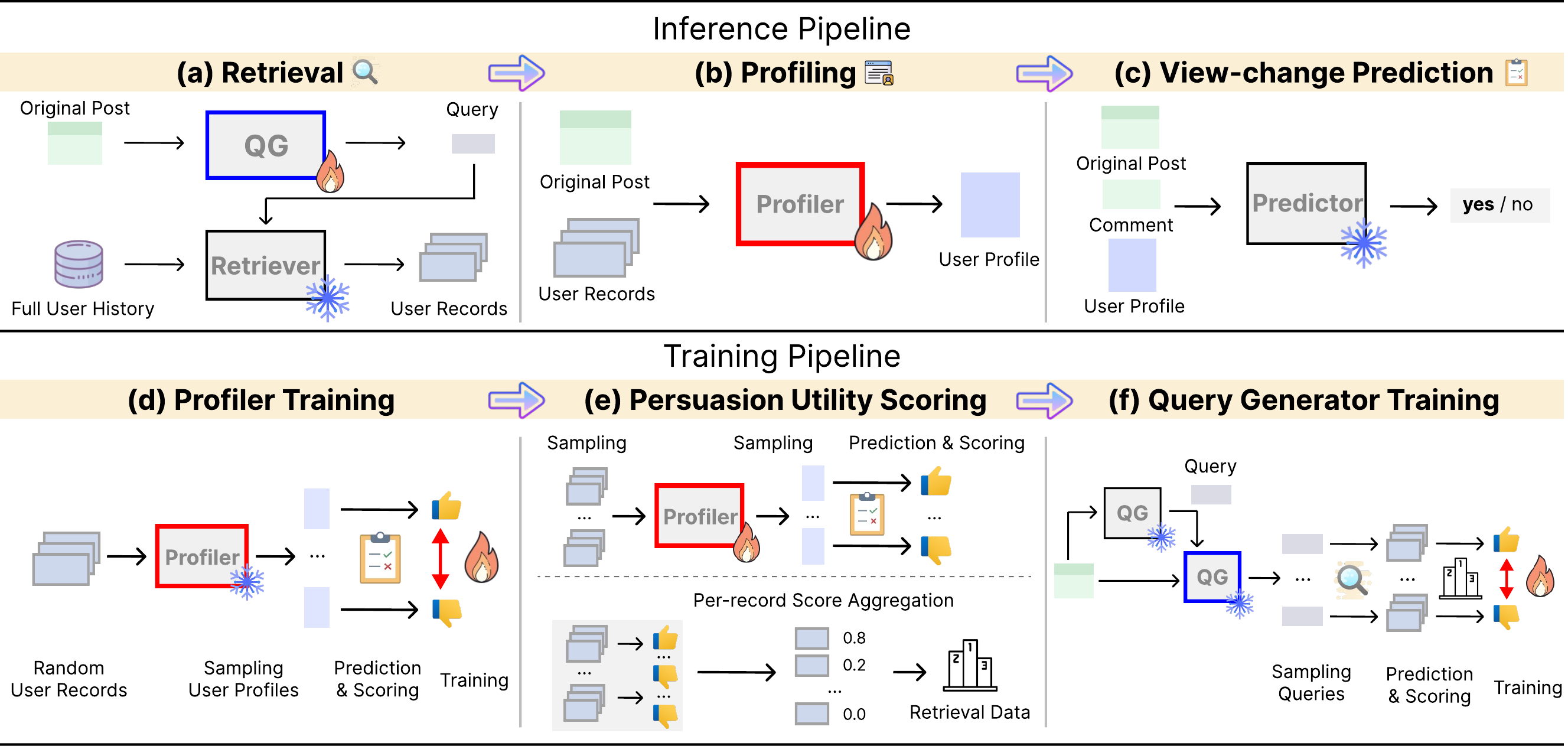}
    \caption{Overview of the proposed framework. \textit{Flame} and \textit{snowflake} icons denote trainable and frozen models, respectively. 
    QG denotes our query generator model.
    The top illustrates the inference pipeline with three stages: (a) retrieval, (b) profiling, and (c) view-change prediction. The bottom shows the training pipeline, consisting of (d) profiler training, (e) record-level persuasion utility scoring, and (f) query generator training.
}
    \label{fig:pipeline}
\end{figure*}

\subsection{Problem Formulation}
\label{sec:problem_formulation}
We construct a dataset from the \texttt{ChangeMyView} (CMV) Reddit forum, where users post opinions and award a \emph{delta} to comments that change their views.
The discussions cover diverse topics, including politics, personal values, and everyday issues.
To support personalized prediction, we collect each user’s historical posts and comments from both CMV and other subreddits.
We split this dataset into training, validation, and test sets with an 8:1:1 ratio, partitioning at the original poster (OP) level so that no OP appears in more than one split. This prevents leakage of OP-specific information from training to evaluation. For training and validation, we subsample up to 100 user records per post.
This is necessary because user records, which are scored via repeated view-change prediction (Section~\ref{sec:persuasion_utility_scoring}), are large and highly variable in size (mean 784, max 19K records per post).
Concretely, we use the \emph{delta} comment as a retrieval query and build the pool using a hybrid retriever that combines BM25 with BGE-M3 semantic similarity.
This substantially reduces computational cost while preserving records that are informative for view-change prediction.

To formalize the task, we represent each data instance as a tuple $(u, x_i, c_i, y_i, R_u)$. 
Here, $u$ denotes a user; $x_i$ is an original post authored by $u$ in the CMV forum that expresses the user’s initial view on a topic; and $c_i$ is a comment written by another user in response to $x_i$. 
Each comment is labeled as a \emph{delta} or \emph{non-delta}, with a label $y_i \in \{0,1\}$ indicating whether it changed the user’s view ($y_i=1$ if a \emph{delta} was awarded).
We further provide access to the user’s historical records $R_u = \{ r^{u,1}, r^{u,2}, \ldots, r^{u,|R_u|} \}$, where each $r^{u,j}$ is a Reddit post or comment written by $u$ prior to $x_i$. 
The personalized view-change task is formulated as $\tilde{y}_i = f(u, R_u, x_i, c_i)$, where the goal is to predict whether $c_i$ will change user $u$’s view expressed in $x_i$, given the user’s history $R_u$.

\subsection{Inference}

To address this task, we introduce a three-stage inference pipeline comprising \emph{retrieval}, \emph{profiling}, and \emph{view-change prediction} (Figure~\ref{fig:pipeline}a--c).
Since $R_u$ is typically large and noisy, direct conditioning on the entire set is impractical.
Instead, we construct a compact user profile $P_i$ that summarizes persuasion-relevant information about $u$ and condition the prediction solely on $P_i$.
The construction of $P_i$ consists of two stages: retrieval and profiling.

The retrieval stage (Figure~\ref{fig:pipeline}a) selects a subset of $k$ records from $R_u$ that are directly used for profile construction. 
For this, we first generate a retrieval query $q_i$ using a trainable query generator $\phi^{\text{query}}$, which takes the original post $x_i$ as input.
Using an embedding-based retriever $\mathcal{M}^{\text{ret}}$, we retrieve the top-$k$ records most relevant to $q_i$: 
\[ \{ r^{u,i_1}, \ldots, r^{u,i_k} \} = \mathcal{M}^{\text{ret}}(q_i, R_u, k) \subseteq R_u. \]
The profiling stage (Figure~\ref{fig:pipeline}b) constructs a natural-language user profile $P_i$ by summarizing the retrieved records into a textual representation that the predictor model can effectively utilize.
We employ a trainable LLM-based profiler $\phi^{\text{prof}}$ that takes the retrieved records and the original post $x_i$ as input, enabling the profile to be conditioned on the persuasion context expressed in $x_i$:
\[P_i = \phi^{\text{prof}}\big( \mathcal{M}^{\text{ret}}(q_i, R_u, k) \, ; \, x_i \big).\]
Finally, at the prediction stage (Figure~\ref{fig:pipeline}c), an LLM-based predictor $\mathcal{M}^{\text{pred}}$ takes post $x_i$, comment $c_i$, and the user profile $P_i$ as input to predict whether $c_i$ will change the user’s view expressed in $x_i$:
\[
\tilde{y}_i = \mathcal{M}^{\text{pred}}(x_i, c_i \, ; \, P_i).
\]

\subsection{Training}

Our framework consists of two learnable components: the query generation module ($\phi^{\text{query}}$) and the profiler ($\phi^{\text{prof}}$).
We jointly optimize them in three stages: \emph{profiler training}, \emph{record-level persuasion utility scoring}, and \emph{query generator training} (Figure~\ref{fig:pipeline}d--f).
We first train the profiler to generate effective user profiles from retrieved records for view-change prediction (Section~\ref{sec:profiler_training}). Next, we train the query generator to produce queries that can retrieve effective user records (Section~\ref{sec:query_generation}), where the utility of each user record is scored based on the effectiveness of its resulting profiles for view-change prediction (Section~\ref{sec:persuasion_utility_scoring}).

\subsubsection{Profiler}
\label{sec:profiler_training}
Our core hypothesis is that an ideal user profile should be optimized for personalized view-change prediction rather than merely summarizing user history.
Since ground-truth profiles for this task are unavailable, we adopt a weakly supervised approach and optimize the profiler via DPO (Figure~\ref{fig:pipeline}d).
We use end-task performance as the preference signal: profiles that successfully predict view change are treated as \emph{chosen}, while unsuccessful ones are treated as \emph{rejected}.
To construct such preference data, for each post $x$, we randomly sample multiple groups of historical records and prompt the base profiler to generate candidate profiles.
We evaluate each profile by its task performance across all comments associated with $x$, using the resulting F1 score as a measure of profile quality.
Preference pairs are then constructed by pairing higher-scoring profiles with lower-scoring ones separated by a sufficient F1 margin, and the profiler is trained to prefer higher-quality profiles.
Training details are provided in Appendix~\ref{sec:app_profiler_details}.

\subsubsection{Record-Level Persuasion Utility Scoring}
\label{sec:persuasion_utility_scoring}

After training the profiler, we estimate the \emph{persuasion utility} of individual user records.
This step enables learning in the retrieval stage: training a record-selection module requires supervision that reflects how useful each record is for predicting persuasion outcomes.
However, existing datasets lack ground-truth annotations identifying which individual records are most informative for persuasion prediction, and collecting such labels from Reddit users is infeasible.
We therefore derive record-level supervision by estimating each record’s contribution to view-change prediction performance (Figure~\ref{fig:pipeline}e).
For each record $r \in R_u$, we estimate its contribution by evaluating its effect across different record sets that include $r$.
Specifically, we randomly partition the records into groups of five, and repeat this grouping process three times.
For each group, we generate three user profiles using the trained profiler with a decoding temperature of $0.7$.
As a result, each record $r$ is associated with a total of nine generated profiles.
We aggregate the F1 scores from all view-change prediction instances performed using profiles that include record $r$, and use the result as its persuasion utility score.

\subsubsection{Query Generator}
\label{sec:query_generation}
For effective user profiling, we must retrieve historical user records informative for persuading each user.
A naive approach would use the post text directly as a retrieval query. However, CMV posts often lack explicit user-specific attributes critical for persuasion—such as underlying values, relevant experiences, or decision-making styles. Post-only queries thus often fail to retrieve sufficiently useful records for predicting view change. To address this, we train an LLM-based query generator to produce user-focused queries that explicitly target attributes absent from the post but potentially critical for persuasion. 

Because training the model to directly infer these implicit attributes from the post alone is difficult, we adopt a two-stage training for the query generator  (Figure~\ref{fig:pipeline}f).
First, we prompt the model to generate a user-focused question that targets information not present in the post but likely to influence persuasion (e.g., for a healthcare policy post: ``What are the user's core values regarding government intervention in individual choice?''). Second, we train the model to take both the post and the generated user-focused question as input and generate a single retrieval query that contextualizes the user attribute using salient cues from the post (e.g., “Does the user prioritize individual autonomy over collective benefit when it comes to healthcare access?”). By learning to ground user attributes in the context of the post, the model can better identify which user information is likely to affect persuasion outcomes.

For each post, we sample multiple candidate queries, retrieve user records for each, and score each query by $\mathrm{NDCG}@5$ based on the persuasion utility of the retrieved records. The model is then trained via DPO to prefer queries that yield higher-quality retrieval. At inference, the trained query generator receives only the post and outputs a user-focused query that effectively surfaces persuasion-relevant user records.
Full details of candidate generation, preference construction, and optimization are provided in Appendix~\ref{app:query_training}.

\section{Experiments}
\label{sec:experiments}

In this section, we evaluate our proposed framework through two complementary analyses:
(1) a retrieval-side evaluation of our query generation strategy based on persuasion utility scores (Section~\ref{sec:results_retrieval});
(2) end-to-end view-change prediction performance, which evaluates the combined effect of all pipeline components (Section~\ref{sec:results_end2end}).

\paragraph{Experimental Setup}
We use data collected from the CMV Reddit forum, as described in Section~\ref{sec:problem_formulation}.
Detailed dataset statistics are provided in Appendix~\ref{sec:data_statistics}.
For the learnable components, we employ \texttt{Llama-3.1-8B-Instruct} as the backbone for both the query generator and the profiler. 
For embedding-based retrieval, we use the \texttt{BGE-M3} embedding model.
For view-change predictor models, we use two open-weight models at different scales,
\texttt{Llama-3.1-8B-Instruct} and \texttt{Llama-3.3-70B-Instruct},
and a closed-source model, \texttt{GPT-4o-mini}, to assess whether our trainable user profiling framework generalizes across predictors.
We evaluate performance using the F1 score, which is well-suited to the inherently imbalanced comment labels (e.g., few delta comments among many non-delta comments).

\begin{table}[t]
\begin{center}
\small
\begin{tabular}{lcc}
\toprule
\textbf{Query Strategy} & \textbf{Mean NCG@5} & \textbf{Mean NDCG@5} \\
\midrule
Random              & 0.6173 & 0.6080 \\
\midrule
BGE-Post            & 0.6267 & 0.6180 \\
BGE-Post-Tuned      & 0.6280 & 0.6162 \\
HyDE                & 0.6229 & 0.6126 \\
\midrule
\textit{Ours}          & \textbf{0.6357} & \textbf{0.6214} \\
\bottomrule
\end{tabular}
\captionof{table}{Retrieval performance of different query formulation strategies.
Random reports the average performance over 10 runs.
BGE-Post and BGE-Post-Tuned use an embedding-based retriever based on BGE-M3, with and without retriever fine-tuning, respectively.}
\label{tab:retrieval_main}
\end{center}
\end{table}

\begin{table*}[t]
\centering
\small
\begin{tabular}{lcccccc}
\toprule
\multirow{2}{*}{Method} 
& \multicolumn{2}{c}{Llama-3.1-8B-Instruct} 
& \multicolumn{2}{c}{Llama-3.3-70B-Instruct} 
& \multicolumn{2}{c}{GPT-4o mini} \\
\cmidrule(lr){2-3} \cmidrule(lr){4-5} \cmidrule(lr){6-7}
& F1 & AUC & F1 & AUC & F1 & AUC \\
\midrule
No Personalization & 0.3457 & 0.5677 & 0.3284 & 0.6538 & 0.2525 & \textbf{0.6415} \\
\midrule
PAG        & 0.2571 & 0.5775 & 0.3141 & 0.6346 & 0.0833 & 0.6165 \\
Recursumm & 0.3133 & 0.5869 & 0.4139 & 0.6571 & 0.1050 & 0.6318 \\
Hsumm     & 0.3244 & 0.5965 & 0.4063 & 0.6615 & 0.1128 & 0.6214 \\
\midrule
Retrieval-only & 0.2952 & 0.5424 & 0.4177 & 0.6635 & 0.1323 & 0.6306 \\
\midrule
\textit{Ours} & \textbf{0.4000} & \textbf{0.6158} & \textbf{0.4661} & \textbf{0.6828} & \textbf{0.2787} & 0.6299 \\
\bottomrule
\end{tabular}
\caption{End-to-end comparison of our proposed framework with prior user profiling approaches.
The table reports F1 and area under the ROC curve (AUC) for view-change prediction across three predictor models.}
\label{tab:e2e_results}
\end{table*}

\subsection{Retrieval-side Experiments}
\label{sec:results_retrieval}

\paragraph{Setup}
We analyze the retrieval component in isolation, focusing on how query generation affects the retrieval of persuasion-relevant user records.
Specifically, we compare a random baseline (mean over 10 runs), embedding-based retrieval methods, and our query generation strategies using pre-computed utility scores for individual records (Section~\ref{sec:persuasion_utility_scoring}).
For embedding-based retrieval baselines, we evaluate 
\textsc{BGE-Post}, which directly use the original post text as the retrieval query, and \textsc{HyDE} \citep{gao-etal-2023-precise}, which generates a hypothetical document that approximates the retrieval target and uses it as the query.
Concretely, for \textsc{HyDE}, we prompt \texttt{Llama-3.1-8B-Instruct} with the original post to generate a plausible user record that is likely to be relevant in the given persuasion context.

\paragraph{Results}

Table~\ref{tab:retrieval_main} reports retrieval performance measured by utility-based NCG@5 and NDCG@5.
The results indicate that dense retrieval using the post text as the query (BGE-Post) is inherently limited, and that fine-tuning the retriever under the same post-only query formulation (BGE-Post-Tuned) leads to only marginal improvements.
This suggests that the bottleneck is not retriever capacity but the incompleteness of the post as a query for eliciting persuasion-relevant user attributes.
In contrast, our method improves over BGE-Post and HyDE by transforming the post into a user-focused query that explicitly targets missing attributes conditioned on the post.
Consistent with this, end-to-end results (Section~\ref{sec:results_end2end}) show that our framework achieves the best view-change prediction performance, highlighting that persuasion-aware query formulation is more beneficial for the full pipeline.

\subsection{End-to-End View-Change Prediction}
\label{sec:results_end2end}

\begin{table*}[t]
\centering
\small
\setlength{\tabcolsep}{6pt}
\begin{tabular}{lccc ccc ccc}
\toprule
& \multicolumn{3}{c}{Llama-3.1-8B-Instruct}
& \multicolumn{3}{c}{Llama-3.3-70B-Instruct}
& \multicolumn{3}{c}{GPT-4o mini} \\
\cmidrule(lr){2-4} \cmidrule(lr){5-7} \cmidrule(lr){8-10}
\textbf{Retrieval} 
& Demograph. & Base & \textit{Ours}
& Demograph. & Base & \textit{Ours}
& Demograph. & Base & \textit{Ours} \\
\midrule
Recent  & 0.3364 & 0.3805 & 0.3951 & 0.3891 & \textbf{0.4058} & 0.4428 & 0.0714 & 0.1629 & 0.2533 \\
Random  & 0.3199 & 0.3758 & 0.3860 & 0.4038 & 0.3979 & 0.4304 & 0.0578 & 0.1516 & 0.2476 \\
BM25    & 0.3286 & 0.3636 & 0.3742     & 0.3905 & 0.3981 & 0.4218 & 0.0720 & 0.1658 & 0.2754 \\
BGE  & 0.3286 & 0.3410 & 0.3554 & \textbf{0.3912} & 0.3799 & 0.4454 & 0.0663 & 0.1465 & 0.2441 \\
HyDE    & 0.3344 & 0.3701 & 0.3785 & 0.3800 & 0.3917 & 0.4507 & 0.0720 & \textbf{0.1805} & 0.2570 \\
\midrule
\textit{Ours} 
& \textbf{0.3466} & \textbf{0.3893} & \textbf{\underline{0.4000}}
& 0.3837 & 0.3929 & \textbf{\underline{0.4661}}
& \textbf{0.0765} & 0.1695 & \textbf{\underline{0.2787}} \\
\bottomrule
\end{tabular}
\caption{Effect of retriever and profiler choices on view-change prediction under different predictors (F1).
Random reports the average performance over 10 random runs.
\underline{Underlined} results denote our final proposed method, while \textbf{boldface} highlights the best-performing configuration within each column.
Column groups correspond to different predictor models, with sub-columns indicating profiler configurations (demographic, base profiler, and our trained profiler).
Corresponding results using the AUC metric are reported in Appendix \ref{appendix_AUC}.}
\label{tab:e2e_results_detailed}
\end{table*}

\paragraph{Setup}

We evaluate the end-to-end view-change prediction performance of our overall pipeline, comparing it against (1) existing personalized profiling frameworks, and (2) different combinations of retrieval and profiling baselines. 

We first compare our method with prior \textbf{user profiling frameworks} for personalized dialogue and retrieval-augmented generation, including \textsc{PAG} \citep{richardson2023integrating}, \textsc{HSumm} \citep{zhong2024memorybank}, and \textsc{Recursumm} \citep{wang2025recursively}.
Details of these baselines are provided in Appendix~\ref{sec:e2e_baselines}.
We additionally evaluate two ablations: \textsc{No Personalization}, which performs view-change prediction without user profiles or historical records, and \textsc{Retrieval-only}, which conditions the predictor on raw retrieved records without profile construction.

Next, we conduct a more detailed comparison across different retriever-profiler combinations.
For \textbf{retrieval} variants, we compare embedding-based strategies evaluated in Section~\ref{sec:results_retrieval} (\textsc{BGE-Post} and \textsc{HyDE}), a sparse retrieval baseline using the post as the query (\textsc{BM25-Post}), and heuristic baselines (\textsc{Random} and \textsc{Recent}).
For \textbf{profiling} variants, we consider three approaches to user profile construction:
(i) \textsc{Demographic}, which extracts demographic attributes from retrieved records using \texttt{GPT-4.1-mini} \citep{hackenburg2025levers, salvi2024conversational},
(ii) \textsc{Base Profiler}, an instruction-tuned LLM without additional training  prompted to summarize retrieved records, and
(iii) \textsc{DPO Profiler}, our profiler trained via DPO (Section~\ref{sec:profiler_training}). \paragraph{Results}
Table~\ref{tab:e2e_results} shows that existing personalization frameworks transfer poorly to view-change prediction.
These methods primarily aim to generate user-aligned responses or compress a user’s history, and generic profiles can even hurt performance for \texttt{Llama-3.1-8B-Instruct} and \texttt{GPT-4o-mini} compared to \textsc{No Personalization}.
In contrast, our framework yields consistent gains across predictors, raising F1 on \texttt{Llama-3.3-70B-Instruct} from $0.3284$ (\textsc{No Personalization}) to $0.4661$, indicating that task-oriented, trainable profiling is crucial for personalized persuasion prediction.
Compared to the retrieval-only baseline, our approach yields consistent gains across all predictors; for example, F1 on \texttt{Llama-3.3-70B-Instruct} improves from $0.4177$ to $0.4661$, highlighting the critical role of the profiler.

Table~\ref{tab:e2e_results_detailed} further decomposes performance by retriever--profiler combinations.
Our DPO-trained profiler consistently outperforms demographic and base profiling baselines across all predictors, while demographic profiles perform poorly, suggesting that persuasion-relevant signals are not well captured by demographics alone.
On the retrieval side, our query generator delivers the strongest end-to-end performance overall; notably, \textsc{Recent} is a competitive baseline, which aligns with \citet{zhang2025prime}. 
Our query generator shows substantial synergy with the trained profiler, highlighting that record-level scoring using the trained profiler provides a clear learning signal.
Together, these results highlight that view-change prediction benefits most from profiles optimized for the task and retrieval queries that expose persuasion-relevant user attributes, rather than from generic personalization pipelines or standard post-only retrieval.

\subsection{Generalization to Other Datasets}
\label{sec:other_datasets}

To test whether our framework generalizes beyond CMV, we evaluate it
on two personalization datasets with different characteristics:
OpinionQA~\citep{santurkar2023whose}, a survey-based stance
prediction task, and PRISM~\citep{kirk2024prism}, a real-world
multi-session conversational preference dataset. These two settings
differ from CMV in both the form of user history and the prediction
target, providing a stringent test of transferability.

For OpinionQA, no interaction history is available; we instead
construct each user's history from their responses to \emph{other}
survey questions, testing whether our framework can leverage
non-conversational user records. Prediction accuracy replaces delta
labels as the supervision signal for utility scoring. For PRISM, we
treat each user's top-rated response as a positive instance and
responses with preference score gap $\geq 0.3$ from the top as
negatives, enabling the same pairwise formulation as CMV. We use \texttt{Llama-3.1-8B-Instruct} as the predictor.

Tables~\ref{tab:other_profiling} and~\ref{tab:other_retrieval} show
that our framework outperforms all profiling and retrieval baselines
on both datasets. The OpinionQA gain is particularly notable: user
history here consists of isolated survey responses rather than
interaction logs, yet our framework still produces effective
profiles. This indicates that the benefit comes from task-oriented
profile construction, not from properties specific to Reddit-style
discussion history. We also observe that full-history summarization
baselines (\textsc{HSumm}, \textsc{RecurSumm}) fall below the No
Personalization baseline on both datasets. This mirrors our CMV
findings and suggests that generic summarization fails to capture
task-relevant user signals across domains.

\begin{table}[t]
\centering
\small
\setlength{\tabcolsep}{4pt}
\begin{tabular}{lcc}
\toprule
\textbf{Profiling} & \textbf{PRISM (F1)} & \textbf{OpinionQA (Acc.)} \\
\midrule
No Personalization & 0.5751 & 0.4354 \\
\textsc{RecurSumm} & 0.5466 & 0.4218 \\
\textsc{HSumm}     & 0.5051 & 0.4014 \\
\textsc{PAG}       & 0.4754 & 0.5238 \\
\textbf{Ours}      & \textbf{0.5879} & \textbf{0.5306} \\
\bottomrule
\end{tabular}
\caption{Profiling comparison on PRISM and OpinionQA with
\texttt{Llama-3.1-8B-Instruct} as the predictor.}
\label{tab:other_profiling}
\end{table}

\begin{table}[t]
\centering
\small
\setlength{\tabcolsep}{4pt}
\begin{tabular}{lcc}
\toprule
\textbf{Retrieval} & \textbf{PRISM (F1)} & \textbf{OpinionQA (Acc.)} \\
\midrule
Random & 0.5686 & 0.4323 \\
BM25   & 0.5678 & 0.4898 \\
BGE    & 0.5814 & 0.4966 \\
\textbf{Ours} & \textbf{0.5879} & \textbf{0.5306} \\
\bottomrule
\end{tabular}
\caption{Retrieval comparison on PRISM and OpinionQA with
\texttt{Llama-3.1-8B-Instruct} as the predictor.}
\label{tab:other_retrieval}
\end{table}

\section{Analysis}
\subsection{Efficiency Analysis}
\label{sec:efficiency}

Since our motivation targets decision-support applications where
inference latency matters, we analyze per-instance token usage and
FLOPs under a realistic deployment scenario in which user history
embeddings are maintained offline and only newly added records
trigger the pipeline; stage-wise breakdowns are in
Appendix~\ref{sec:appendix_efficiency}.

Table~\ref{tab:efficiency} reveals a clear gap between methods that
consume the entire user history and those that selectively retrieve.
\textsc{HSumm} and \textsc{RecurSumm} must reconstruct the user
profile from the full history whenever a new input arrives,
incurring $6$--$13\times$ our cost in both tokens and FLOPs. This
overhead grows linearly with user history size, making such
approaches increasingly impractical for long-lived users.
Retrieval-based methods (\textsc{PAG}, ours) instead scale with the
retrieved subset. Compared to \textsc{PAG}, our query-generation
step adds only $\sim$$6\%$ tokens and $\sim$$8\%$ FLOPs---a modest
overhead against our average $+0.16$ F1 gain across predictors
(Table~\ref{tab:e2e_results}). Per-inference cost is thus bounded
by the retrieved subset rather than accumulated history, aligning
with the real-time decision-support setting that motivates our work.

\begin{table}[t]
\centering
\small
\setlength{\tabcolsep}{4pt}
\begin{tabular}{lrr}
\toprule
\textbf{Method} & \textbf{Tokens} & \textbf{FLOPs} \\
\midrule
\textsc{PAG}       & 3{,}228  & $4.71\times10^{13}$ \\
\textbf{Ours}      & 3{,}429  & $5.09\times10^{13}$ \\
\textsc{RecurSumm} & 21{,}678 & $3.58\times10^{14}$ \\
\textsc{HSumm}     & 42{,}880 & $7.03\times10^{14}$ \\
\bottomrule
\end{tabular}
\caption{Per-instance inference cost across profiling methods.}
\label{tab:efficiency}
\end{table}

We additionally note that record-level persuasion utility scoring
(Section~\ref{sec:persuasion_utility_scoring}) is the most expensive training
component, consuming approximately $1.4$B tokens on the training
set with \texttt{GPT-4o-mini}. Crucially, however, this cost is
incurred \emph{only once} during training and does not affect
inference: once records are scored, the framework applies to new
users without additional training, amortizing the upfront expense
over all downstream inferences.

\subsection{Profiler Analysis}
\label{sec:profiler_analysis}

In this section, we analyze the impact of profiler training by comparing profiles generated by the base profiler (\emph{original profiles}) and the trained profiler (\emph{trained profiles}) on the test set.
We present two key analyses below, with results for all predictor models reported in Appendix~\ref{sec:profiler_analysis_details}.

\paragraph{The effectiveness of profiler training varies by post topic.}

\begin{figure}[t]
    \centering
    \includegraphics[width=1.0\linewidth]{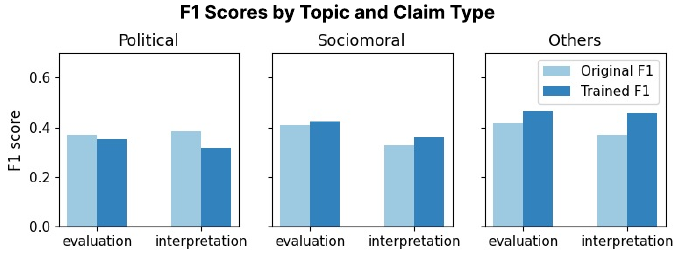}
    \caption{F1 by topic and claim type of the post, comparing the original and trained profilers.
\texttt{Llama-3.1-8B-Instruct} is used as the predictor model.
}
    \label{fig:profiler-analysis-1}
\end{figure}

To analyze how profiler effectiveness varies across post characteristics, we annotate each post by \emph{topic} and \emph{claim type} using \texttt{GPT-4.1-mini}.
Topics are categorized into \emph{Political} (27.4\%), \emph{Sociomoral} (46.4\%), and \emph{Others} (26.2\%), and claim types into \emph{Interpretation} (39.9\%) and \emph{Evaluation} (60.1\%), following prior work~\citep{hidey2017analyzing,priniski2018attitude}.
Across most topic-claim combinations, trained profiles consistently outperform original profiles in F1 (Figure~\ref{fig:profiler-analysis-1}).
The only exception is political posts under \texttt{Llama-3.1-8B-Instruct}, where profiling benefits sociomoral and other topics but not political posts, likely due to the dominance of group identities in political persuasion.
Overall, the results suggest that the trained profiler effectively captures individual-level characteristics relevant to persuasion.

\begin{figure}[t]
    \centering
    \includegraphics[width=\linewidth]{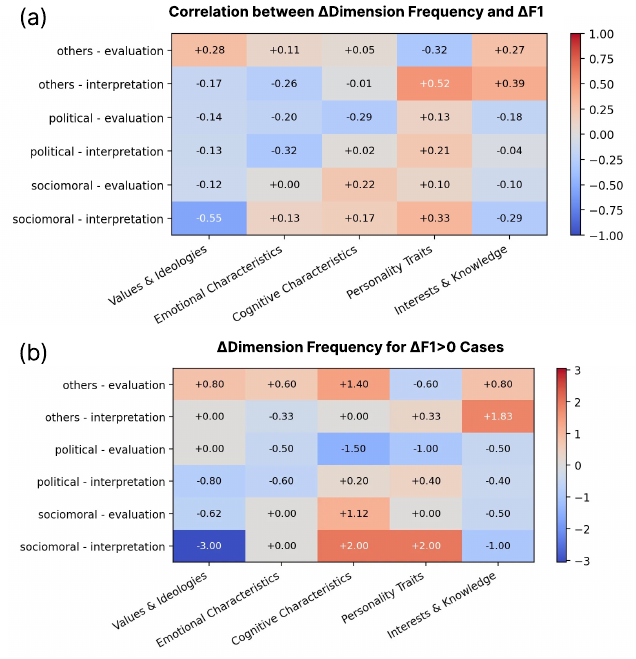}
    \caption{
Analysis of profile-dimension frequency shifts ($\Delta$DF) and performance gains ($\Delta$F1) between the original and trained profilers.
(a) Correlation between $\Delta$DF and $\Delta$F1.
(b) $\Delta$DF for cases with $\Delta$F1 $>$ 0.
\texttt{Llama-3.1-8B-Instruct} is used as the predictor model.
}
    \label{fig:profiler-analysis-2}
\end{figure}

\paragraph{Important profile dimensions vary by post characteristics and the predictor model.}

To further analyze profile content, we decompose each profile into sentence-level units, referred to as \emph{profile items}.
We annotate each item for the presence of five profile dimensions using \texttt{GPT-4.1-mini}; the five dimensions---\emph{Values \& Ideologies}, \emph{Emotional Characteristics}, \emph{Cognitive Characteristics}, \emph{Personality Traits}, and \emph{Interests \& Knowledge}---are constructed from persuasion literature \citep{fabrigar1999role, al2020exploiting}.
For each profile, we count the frequency of profile items associated with each dimension and compute \emph{Profile-F1}, the F1 score aggregated over all comments associated with the profile.
Figure~\ref{fig:profiler-analysis-2}a shows the correlation between (i) change in item frequency for each dimension from original to trained profiles and (ii) the change in Profile-F1.

Our analysis yields three key findings:
(1) No single profile dimension is consistently beneficial or detrimental across all posts.
(2) The effect of each dimension is strongly post-dependent: for example, cognitive traits (e.g., reasoning styles, decision making styles) are positively associated with performance gains for political evaluation posts, but negatively associated with sociomoral evaluation posts.
(3) For cases where performance improves, shifts in item frequency across profile dimensions align with these correlation trends (Figure~\ref{fig:profiler-analysis-2}b), indicating that different dimensions are emphasized depending on the post.
These three patterns remain consistent across different predictor models.
However, the specific association patterns between post characteristics and profile dimensions vary substantially with the choice of predictor model.
Taken together, these results suggest that persuasion-relevant dimensions differ across posts and the predictor models, and that the trained profiler captures this post-dependent and predictor-specific variation by dynamically adjusting the emphasis on different dimensions.
Results for all predictor models are reported in Appendix~\ref{sec:profiler_analysis_details}.

\subsection{User Record Analysis}

We conduct an analysis of user records scored via persuasion-utility scoring, using \textsc{Llama-3.1-8B-Insruct} as the predictor model, focusing on (1) semantic differences between high- and low-scoring records and (2) cross-model patterns in utility scores. More detailed analyses are provided in Appendix~\ref{sec:record-analysis}.

\paragraph{Low-scoring records are not semantically dissimilar to the post.}
We analyze pairs of top-1 and bottom-1 records for the same post in the validation set, focusing on cases where the bottom-1 record receives an F1 score of zero. Following Section~\ref{sec:profiler_analysis}, we annotate the records along \emph{topic} and \emph{claim type}. Contrary to the hypothesis that low-scoring records fail due to semantic misalignment with the post, we observe the opposite trend:
low-scoring records are more likely than high-scoring ones to share the same topic or claim type as the post.
This highlights the need for finer-grained contextualization in persuasion.

\paragraph{Different predictor models prefer different user records.}

We further compare persuasion utility scores across predictor models and find little agreement in their preferred records (Table~\ref{tab:pairwise_agreement_score}).
Pairwise comparisons of the top-5 records show low overlap (0.24--0.28), corresponding to only about 1.25 shared records on average.
Similarly, Spearman rank correlations are near zero (-0.005--0.083), indicating weak consistency in relative ordering.
These results indicate that record utility for view-change prediction is highly model-dependent, motivating the training of predictor-specific retrieval modules.


\begin{table}[t]
\centering
\small
\begin{tabular}{lcc}
\toprule
\textbf{Model Pair} 
& \textbf{Top-5 Overlap} 
& \textbf{Spearman $\rho$} \\
\midrule
GPT / Llama70B 
& 0.273 
& 0.007 \\

GPT / Llama8B 
& 0.281 
& 0.083 \\

Llama70B / Llama8B 
& 0.245 
& $-0.005$ \\
\bottomrule
\end{tabular}
\caption{
Pairwise agreement between predictors on record-level utility scores, measured by mean top-5 overlap and Spearman $\rho$.
GPT, Llama8B, and Llama70B correspond to \texttt{GPT-4o-mini}, \texttt{Llama-3.1-8B-Instruct}, and \texttt{Llama-3.3-70B-Instruct}, respectively.
}
\label{tab:pairwise_agreement_score}
\end{table}

\section{Conclusion}

We introduce a trainable user profiling framework that captures persuasion-relevant user factors.
Experiments on the CMV dataset show that our approach consistently outperforms baselines by constructing context-dependent profiles tailored to the downstream predictor, with consistent gains also observed on OpinionQA and PRISM.
By learning to retrieve and construct task-oriented user profiles, our framework enables scalable, context-sensitive personalization without retraining predictors or requiring extensive user annotations, at 6--13× lower inference cost than full-history summarization baselines, making it practical for real-world decision-support systems such as conversational agents, recommendation, and coaching.




\section*{Limitations}
This study focuses on personalized persuasiveness prediction in the setting of online opinion change, evaluated on the \texttt{ChangeMyView} Reddit dataset. While this setting provides a well-established testbed with explicit view-change signals, it represents a specific form of persuasion grounded in long-form textual discussions. Extending the framework to other interaction modalities or domains---such as short-form conversations or real-time recommendation settings---would require additional validation.

\section*{Ethical Considerations}

Research on predicting view change in online discussions could be related to ethical considerations about user autonomy and the responsible use of predictive insights. In this work, we strictly focus on predicting whether a view change occurs, rather than intervening in user behavior. The framework does not include any mechanisms for generating persuasive content or intervening on individuals, but is designed to enhance understanding of view-change dynamics in natural settings.

All experiments are conducted using publicly available and anonymized data, without any personally identifiable information.  

\section*{Acknowledgement}
This work was supported by the Creative-Pioneering Researchers Program through Seoul National University and by the National Research Foundation of Korea (NRF) grants (RS-2024-00333484 and RS-2024-00414981) funded by the Korean government (MSIT).

\section*{AI Assistance Acknowledgement}
We used AI assistants to proofread the writing and to help with coding.





\bibliography{custom}

\appendix

\newpage
\section{Dataset Collection and Statistics}
\label{sec:data_statistics}
We construct our dataset from the \texttt{ChangeMyView} (CMV) Reddit forum. To ensure the dataset supports the study of personalized persuasion with sufficient historical context, we applied a multi-stage filtering process to the raw data:

\begin{enumerate}
    \item \textbf{Interaction Completeness:} We filtered for original posts (OPs) that contain at least one \textit{delta} comment (successful persuasion) and at least one \textit{non-delta} comment (unsuccessful persuasion). This ensures that each instance allows for the comparison of successful and unsuccessful arguments within the same context.
    \item \textbf{History Availability:} We restricted the dataset to users who had at least 15 historical records (posts or comments) published prior to the timestamp of the original post. This threshold ensures there is sufficient data to construct a meaningful user profile.
\end{enumerate}

After filtering, the final dataset consists of 1,676 posts from 1,326 unique users. We split this dataset into training, validation, and test sets with an approximate ratio of 8:1:1. Table~\ref{tab:dataset_stats} presents the detailed statistics for each split, including the distribution of user history length and the volume of delta and non-delta comments.

\begin{table*}[h]
\centering
\resizebox{\textwidth}{!}{%
\begin{tabular}{lrrrrrrrrrrrr}
\toprule
 & & & \multicolumn{4}{c}{\textbf{User History Count}} & \multicolumn{3}{c}{\textbf{Delta Comments}} & \multicolumn{3}{c}{\textbf{Non-Delta Comments}} \\
\cmidrule(lr){4-7} \cmidrule(lr){8-10} \cmidrule(lr){11-13}
\textbf{Split} & \textbf{\# Posts} & \textbf{Unique OPs} & \textbf{Min} & \textbf{Max} & \textbf{Mean} & \textbf{Median} & \textbf{Min} & \textbf{Mean} & \textbf{Median} & \textbf{Min} & \textbf{Mean} & \textbf{Median} \\
\midrule
Train & 1,341 & 1,257 & 15 & 11,965 & 252.40 & 57 & 1 & 1.77 & 1 & 1 & 33.06 & 20.0 \\
Validation & 167 & 69 & 16 & 19,583 & 956.35 & 65 & 1 & 2.24 & 1 & 1 & 31.81 & 19.0 \\
Test & 168 & 69 & 15 & 19,583 & 613.36 & 71 & 1 & 2.54 & 1.5 & 2 & 35.30 & 19.0 \\
\bottomrule
\end{tabular}%
}
\caption{Detailed statistics of the dataset splits. User History Count refers to the number of historical posts/comments available for the OP prior to the current post.}
\label{tab:dataset_stats}
\end{table*}

\section{Predictor Model Prompts}
\label{app:predictor_prompts}

In this section, we provide the detailed prompts used for the predictor models. We present the System Prompt and User Prompt sequentially for each setting.

\subsection{Prediction with User Profile (Ours)}

\noindent\textbf{System Prompt}
\begin{quote}
\small\ttfamily
    \texttt{You are the author of the post. The section labeled "User Profile" is your profile — it describes who you are.} \par\vspace{0.5em}
    \texttt{Read it carefully and fully adopt this as your identity and mindset.} \par\vspace{0.5em}
    \texttt{You will then be shown a post you wrote, and a comment written in response to it. Based on your profile, determine whether the comment would change your mind from the opinion expressed in the post.} \par\vspace{0.5em}
    \texttt{Respond only with one word: "yes" if your mind would change after reading the comment, or "no" if not. Do not provide any explanation or reasoning.}
\end{quote}

\noindent\textbf{User Prompt}
\begin{quote}
\small\ttfamily
    \texttt{\#\#\# User Profile} \par
    \texttt{\{user\_profile\}} \par\vspace{1em}
    \texttt{\#\#\# Post} \par
    \texttt{\{post\}} \par\vspace{1em}
    \texttt{\#\#\# Comment} \par
    \texttt{\{comment\}} \par\vspace{1em}
    \texttt{---} \par
    \texttt{Would this comment change your mind from the opinion you expressed in the post?} \par\vspace{0.5em}
    \texttt{Respond only with one word: "yes" or "no".}
\end{quote}

\subsection{Prediction with User History (Retrieval-Only)}

\noindent\textbf{System Prompt}
\begin{quote}
\small\ttfamily
    \texttt{You are the author of the post. The section labeled "User History" is relevant past history about you.} \par\vspace{0.5em}
    \texttt{Read it carefully and incorporate it into your identity and mindset.} \par\vspace{0.5em}
    \texttt{You will then be shown a post you wrote, and a comment written in response to it. Based on your history, determine whether the comment would change your mind from the opinion expressed in the post.} \par\vspace{0.5em}
    \texttt{Respond only with one word: "yes" if your mind would change after reading the comment, or "no" if not. Do not provide any explanation or reasoning.}
\end{quote}

\noindent\textbf{User Prompt}
\begin{quote}
\small\ttfamily
    \texttt{\#\#\# User History} \par
    \texttt{\{user\_profile\}} \par\vspace{1em}
    \texttt{\#\#\# Post} \par
    \texttt{\{post\}} \par\vspace{1em}
    \texttt{\#\#\# Comment} \par
    \texttt{\{comment\}} \par\vspace{1em}
    \texttt{---} \par
    \texttt{Would this comment change your mind from the opinion you expressed in the post?} \par\vspace{0.5em}
    \texttt{Respond only with one word: "yes" or "no".}
\end{quote}

\subsection{Prediction without Personalization}

\noindent\textbf{System Prompt}
\begin{quote}
\small\ttfamily
    \texttt{You are the author of the post. Carefully read your own post and the comment written in response to it.} \par\vspace{0.5em}
    \texttt{Decide whether you would change your mind after reading the comment.} \par\vspace{0.5em}
    \texttt{Ignore your own beliefs as a language model and fully adopt the mindset of the person who wrote the post.} \par\vspace{0.5em}
    \texttt{Respond with only one word: "yes" if you think you would change your mind, or "no" if not. Do not provide any explanation or reasoning.}
\end{quote}

\noindent\textbf{User Prompt}
\begin{quote}
\small\ttfamily
    \texttt{[Post]} \par
    \texttt{\{post\}} \par\vspace{1em}
    \texttt{[Comment]} \par
    \texttt{\{comment\}} \par\vspace{1em}
    \texttt{Would you change your mind after reading the comment?}
\end{quote}
\section{Profiler Training Details}
\label{sec:app_profiler_details}

In this section, we provide detailed specifications for the preference construction process and the hyperparameters used for Direct Preference Optimization (DPO).

\subsection{Preference Pair Construction}
\label{ssec:preference_construction}
To derive robust training signals from the synthesized candidate profiles, we employ a margin-based stratified sampling strategy. As described in Section~\ref{sec:profiler_training}, for each input group $\mathcal{G}_i$, we generate a set of 16 candidate profiles $\Pi_{\mathcal{G}_i}$. We rank these profiles based on their utility score $S(\pi)$, which represents the macro-F1 score on the view-change prediction task.

To avoid noisy training signals arising from pairs with negligible performance differences, we enforce a minimum utility margin $\delta$. We construct a dataset of preference pairs $\mathcal{D} = \{ (x, \pi_w, \pi_l) \}$ where:
\begin{equation}
    S(\pi_w) - S(\pi_l) \geq \delta
\end{equation}
where $x$ represents the input historical records. Specifically, we select the top-$K$ performing profiles as positive samples and the bottom-$K$ profiles as negative samples from the candidate set $\Pi_{\mathcal{G}_i}$. We then form pairs from the Cartesian product of these two subsets, filtering out any pairs that do not satisfy the margin condition. In our experiments, we set $K=4$ (top 25\% and bottom 25\%) and the margin $\delta = 0.05$ to ensure distinct quality separation.

\subsection{DPO Training Configuration}
\label{ssec:dpo_config}
We optimize the profiler $\pi_\theta$ using the standard DPO objective, which increases the likelihood of the preferred profile $\pi_w$ while decreasing that of the dispreferred profile $\pi_l$, implicitly optimizing the reward function without a separate reward model training step. The loss function is defined as:

\begin{equation}
\begin{aligned}
    \mathcal{L}_{\text{DPO}} & (\pi_\theta; \pi_{\text{ref}}) = \\
    - \mathbb{E} & _{(x, \pi_w, \pi_l) \sim \mathcal{D}} \Big[ \log \sigma \Big( \beta \log \frac{\pi_\theta(\pi_w|x)}{\pi_{\text{ref}}(\pi_w|x)} \\
    & - \beta \log \frac{\pi_\theta(\pi_l|x)}{\pi_{\text{ref}}(\pi_l|x)} \Big) \Big]
\end{aligned}
\end{equation}

where $\pi_{\text{ref}}$ is the frozen reference model (the initial base profiler), $\sigma$ is the logistic sigmoid function, and $\beta$ is a hyperparameter controlling the deviation from the reference model.

We initialized the profiler with \texttt{Llama-3.1-8B-Instruct}. To ensure training stability and prevent overfitting to the small number of high-utility patterns, we utilized Low-Rank Adaptation (LoRA) for parameter-efficient fine-tuning. The detailed hyperparameters are listed in Table~\ref{tab:dpo_hyperparams}.

\begin{table}[h]
    \centering
    \small
    \begin{tabular}{l|c}
    \toprule
    \textbf{Hyperparameter} & \textbf{Value} \\
    \midrule
    Base Model & Llama-3.1-8B-Instruct \\
    LoRA Rank ($r$) & 32 \\
    LoRA Alpha ($\alpha$) & 64 \\
    \midrule
    Optimizer & AdamW \\
    Learning Rate & 5e-7 \\
    LR Scheduler & Linear \\
    Warmup Ratio & 0.05 \\
    Batch Size & 64 \\
    Beta ($\beta$) & 0.1 \\
    Epochs & 3 \\
    Max Sequence Length & 16384 \\
    \bottomrule
    \end{tabular}
    \caption{DPO training hyperparameters for the profiler.}
        \label{tab:dpo_hyperparams}
\end{table}
\FloatBarrier
\subsection{Profile Generation Prompts}
\label{app:profile_gen_prompts}

We use the following prompts to generate a context-aware user profile tailored for persuasion.

\noindent\textbf{System Prompt}
\begin{quote}
\small\ttfamily
    \texttt{You are an expert assistant whose task is to extract concise, high-level information about the author of a set of passages.} \par\vspace{0.5em}
    \texttt{Focus only on traits that would be most useful for persuading or changing the author's view in relation to the current post.} \par\vspace{0.5em}
    \texttt{Your goal is to produce a compact, context-aware user profile optimized for persuasive messaging toward the given post.}
\end{quote}

\noindent\textbf{User Prompt}
\begin{quote}
\small\ttfamily
    \texttt{You are given a set of passages written by the same author, along with the author's current post.} \par\vspace{0.5em}
    \texttt{Extract only the most essential information about the author that is clearly stated or strongly and consistently implied across multiple passages, focusing on traits that are most relevant for understanding how to persuade them in the context of the current post.} \par\vspace{1em}
    
    \texttt{Instructions:} \par
    \texttt{\ \ - Consider the current post as the immediate context in which persuasion would occur.} \par
    \texttt{\ \ - Identify attitudes, reasoning patterns, or sensitivities that could influence how the author might respond to persuasion regarding the post.} \par
    \texttt{\ \ - Do not guess or speculate beyond what is well supported.} \par
    \texttt{\ \ - Exclude personally identifying or sensitive details unless explicitly stated.} \par
    \texttt{\ \ - Generalize from specific events or examples into higher-level traits; avoid direct quotes or low-level details.} \par
    \texttt{\ \ - Remove redundancy and keep bullets concise.} \par
    \texttt{\ \ - Do NOT respond with anything other than the bullet points.} \par\vspace{1em}
    
    \texttt{Current Post:} \par
    \texttt{\{post\}} \par\vspace{1em}
    
    \texttt{Input Passages:} \par
    \texttt{\{passages\}} \par\vspace{1em}
    
    \texttt{Output:} \par
    \texttt{\ \ • ...} \par
    \texttt{\ \ • ...}
\end{quote}

\section{Query Generator Training Details}
\label{app:query_training}

This appendix provides implementation details for training the query generator, including candidate generation, retrieval-based supervision, preference construction, and optimization settings.
Across all experiments, the query generator is implemented as a single LLM (Llama-3.1-8B-Instruct) and trained using Direct Preference Optimization (DPO) with a two-stage training strategy, following Section~\ref{sec:query_generation}.

\subsection{Overview}

The query generator is trained to produce \emph{user-focused retrieval queries} that retrieve historical user records informative for personalized persuasion.
A post-only query is often insufficient because the post may not explicitly mention persuasion-critical user attributes (e.g., values, experiences, decision-making styles).
To make learning easier, we adopt a two-stage training strategy:
\begin{enumerate}
    \item \textbf{Stage 1 (User-Focused Question Generation).} Prompt the model to produce a user-focused \emph{question} that asks for user information \emph{not present in the post} but likely to affect persuasion.
    \item \textbf{Stage 2 (Post-Contextualized Query Generation).} Train the model to take the post and the Stage-1 question as input and generate a single \emph{retrieval query} that contextualizes the user attribute using salient post cues (topic, stance, constraints).
\end{enumerate}

In second stage, supervision is derived from retrieval quality: we score candidate queries by $\mathrm{NDCG}@5$ based on the persuasion utility of retrieved records and apply DPO to prefer candidates with higher retrieval quality.
At inference, the trained model receives only the post and outputs a user-focused retrieval query.

\subsection{Candidate Generation Procedure}

For each post $x_i$, we generate candidates as follows.

\paragraph{Stage 1: User-Focused Question.}
We first generate a single user-focused question $q_i^{(1)}$ from the query generator using the Stage-1 prompt with decoding temperature $0$.
This question serves as an intermediate representation of the user attribute to seek.

\paragraph{Stage 2: Post-Contextualized Retrieval Query.}
Conditioned on $(x_i, q_i^{(1)})$, we sample $16$ candidate retrieval queries $\{q_{i,j}^{(2)}\}_{j=1}^{16}$ using the Stage-2 prompt with temperature $0.8$.
Each candidate is a single natural-language sentence that integrates (i) the user attribute targeted by $q_i^{(1)}$ and (ii) salient cues from $x_i$.

This two-step candidate generation is used to construct DPO training data and is applied consistently across all predictor models.

\subsection{Retrieval and Scoring}

Each Stage-2 candidate query $q_{i,j}^{(2)}$ is used to retrieve the top-$5$ user records from the author’s historical records using a fixed embedding-based retriever (BGE-M3).
We evaluate query quality using $\mathrm{NDCG}@5$, where the graded relevance of each retrieved record is given by its pre-computed \emph{record-level persuasion utility score} (Section~\ref{sec:persuasion_utility_scoring}).

Both the retriever and the utility scores are kept fixed throughout query generator training.
Thus, the query generator is trained solely to improve retrieval quality under a fixed downstream evaluation signal.

\subsection{Preference Pair Construction}

For each post, we partition the $16$ Stage-2 candidates into positive and negative pools based on their $\mathrm{NDCG}@5$ scores, and construct preference pairs by pairing positives with negatives.
We additionally enforce a minimum margin of $0.10$ between the chosen and rejected query scores, and select up to a fixed maximum number of pairs per post.

Because the $\mathrm{NDCG}@5$ score distributions differ across predictor models (due to predictor-specific persuasion utility scoring), we use predictor-specific thresholds and pair caps to ensure sufficient supervision.
Table~\ref{tab:query_pref_thresholds} summarizes the settings.

\begin{table}[t]
\centering
\small
\setlength{\tabcolsep}{4pt}
\begin{tabular}{lccc}
\hline
\textbf{Predictor} & \textbf{Pos.} & \textbf{Neg.} & \textbf{Max} \\
\hline
Llama-3.1-8B-Instruct & $\geq$ 0.65 & $\leq$ 0.55 & 8 \\
Llama-3.3-70B-Instruct & $\geq$ 0.75 & $\leq$ 0.65 & 8 \\
GPT-4o-mini & $\geq$ 0.55 & $\leq$ 0.45 & 10 \\
\hline
\end{tabular}
\caption{Predictor-specific thresholds for preference pair construction based on $\mathrm{NDCG}@5$.}
\label{tab:query_pref_thresholds}
\end{table}

\subsection{Optimization via DPO}

We train the query generator using DPO with LoRA fine-tuning.
All settings are shared across predictors except the DPO inverse temperature $\beta$, which we tune to account for differences in preference sharpness under each predictor’s utility signal.

\begin{table}[t]
\centering
\small
\setlength{\tabcolsep}{6pt}
\begin{tabular}{l l}
\hline
\textbf{Hyperparameter} & \textbf{Value} \\
\hline
Base model & Llama-3.1-8B-Instruct \\
LoRA rank $r$ & 16 \\
LoRA scaling $\alpha$ & 32 \\
Learning rate & $2 \times 10^{-5}$ \\
Max epochs & 3 \\
DPO $\beta$ & 0.3 (0.1 for GPT-4o-mini) \\
\hline
\end{tabular}
\caption{DPO training hyperparameters for the query generator.}
\label{tab:query_dpo_hparams}
\end{table}

\subsection{Query Generator Prompts}
\label{app:query_prompts}

\paragraph{User-Focused Question Prompt.}

\paragraph{System Prompt}

\begin{quote}
\small\ttfamily
You will be given an online post where a user explains their view on a specific topic.\\
Write ONE short question that asks for information regarding the user that is NOT explicitly stated in the post, but would be important for persuading the user expressed in the post.\\
The question should focus on aspects such as the user's values, experiences, priorities, or decision making styles related to the topic.\\[0.5em]
Instructions:
\begin{itemize}
  \item Output MUST be a single question sentence ending with ``?''.
  \item Do NOT explain your reasoning.
  \item Do NOT ask for information already provided in the post.
\end{itemize}
\end{quote}

\noindent\textbf{User Prompt}
\begin{quote}
\small\ttfamily
Post:\\
---\\
\{post\}\\
---\\
Respond in ONE question.
\end{quote}

\paragraph{Post-Contextualized Query Prompt.}

\paragraph{System Prompt}

\begin{quote}
\small\ttfamily
You will be given two inputs:\\
(1) an online post where a user explains their view on a specific topic.\\
(2) a question asking for information that is NOT explicitly stated in the post, but is important for persuading the user in this situation.\\[0.5em]
Write ONE sentence that incorporates:
\begin{itemize}
  \item what the question is asking about the user
  \item the most important cues from the post
\end{itemize}
The sentence should clearly reflect what the question asks about the user, while also grounding it in the most important cues from the post.\\[0.5em]
Instructions:
\begin{itemize}
  \item Output MUST be a single sentence.
  \item Do NOT explain your reasoning.
\end{itemize}
\end{quote}

\noindent\textbf{User Prompt}
\begin{quote}
\small\ttfamily
Post:\\
---\\
\{post\}\\
---\\
Question:\\
---\\
\{question\}\\
---\\
Respond in ONE sentence.
\end{quote}

\section{Examples of the Generated User Profiles}
\label{sec:profile-examples}

\begin{figure}[t]
    \centering
    \includegraphics[width=1.0\linewidth]{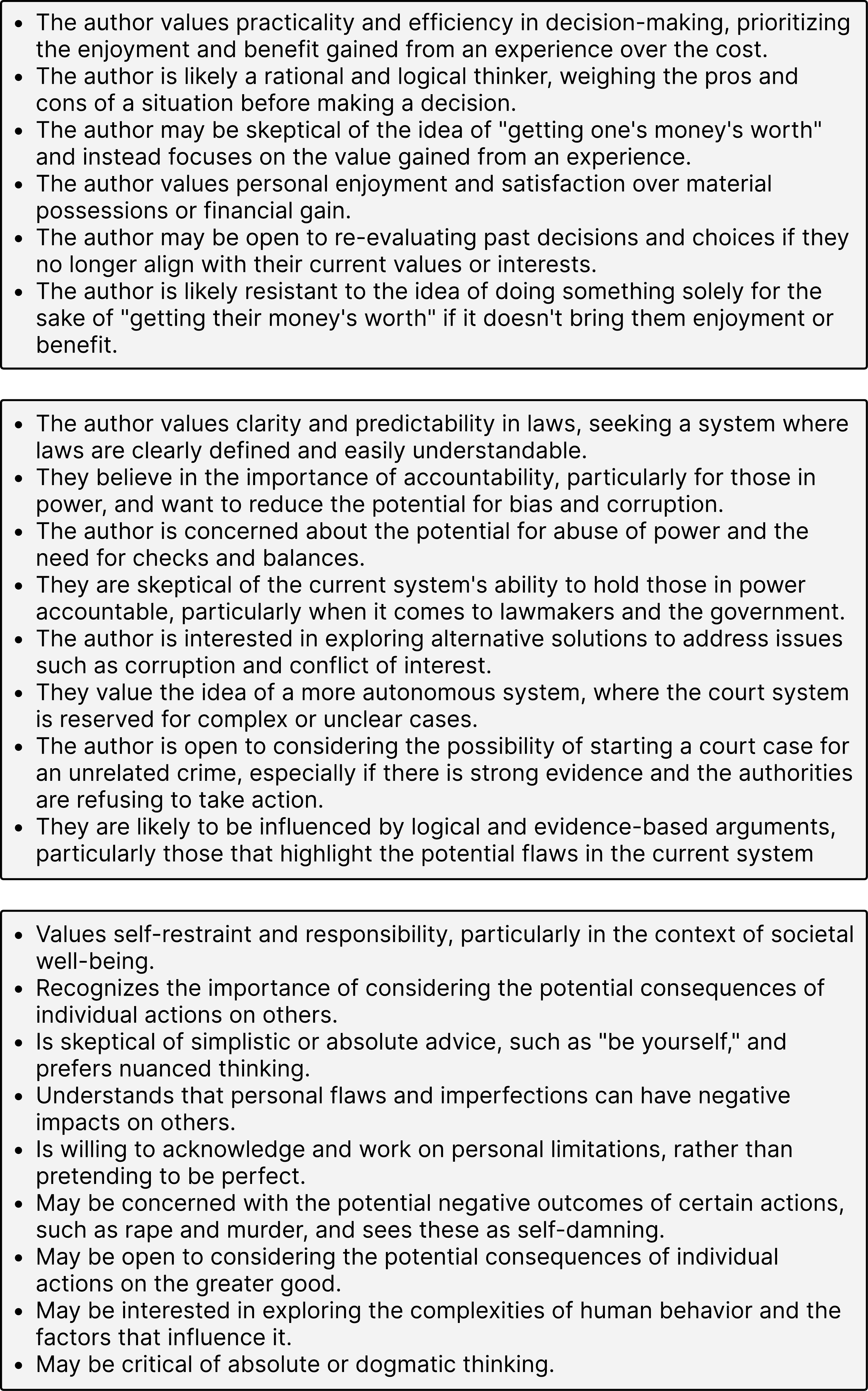}
    \caption{Examples of user profiles generated by our trained profiler when used \texttt{Llama3.1 8B Instruct} as the predictor model for training.
}
    \label{fig:profiler-examples}
\end{figure}

To provide qualitative insight into the behavior of our trained profiler, we present examples of generated user profiles at Figure~\ref{fig:profiler-examples}.

\begin{figure}[t]
    \centering
    \includegraphics[width=1.0\linewidth]{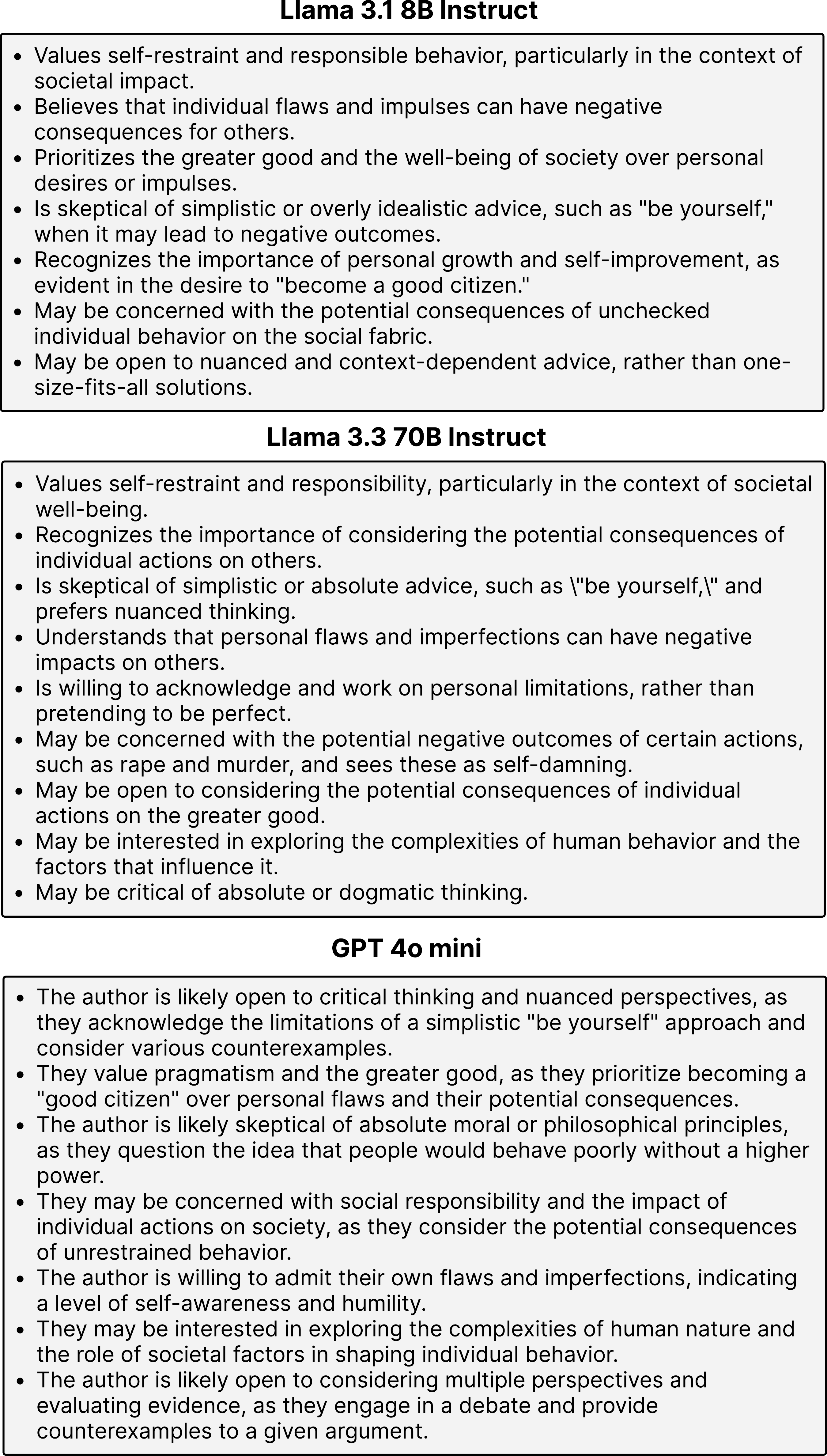}
\caption{Examples of user profiles generated by our trained profilers for the same user under different predictor models used for training.}
    \label{fig:profiler-examples-models}
\end{figure}

Figure~\ref{fig:profiler-examples-models} compares profiles generated for the same user when trained with different predictor models.
Despite being grounded in identical user records, the resulting profiles exhibit systematic differences in emphasis and framing, reflecting predictor-specific preferences for information that is most useful for downstream prediction.
This comparison highlights that our profiler does learn to adapt profile construction tailored to the target predictor model.

\section{Details of User Profiling Baselines}
\label{sec:e2e_baselines}

We compare our approach against several representative user profiling frameworks.
Specifically, we consider:
(i) \textsc{PAG} \citep{richardson2023integrating}, which retrieves a subset of user records using BM25, independently summarizes each selected record, and concatenates the resulting summaries into a user profile;
(ii) \textsc{HSumm} \citep{zhong2024memorybank}, which applies hierarchical summarization by first generating summaries over subsets of user records and then aggregating these intermediate summaries into a single profile;
(iii) \textsc{Recursumm} \citep{wang2025recursively}, which incrementally updates the user profile by recursively integrating each new user record with the existing summary.
\textsc{PAG} relies on retrieval to select a subset of user records for profiling, whereas \textsc{HSumm}, \textsc{Recursumm}, and \textsc{PRIME} assume access to and utilize all available user historical records when constructing user profiles.
Concretely, \textsc{HSumm} and \textsc{Recursumm} construct a profile for each user by repeatedly summarizing that user’s entire history through multiple summarization steps.

All baselines were minimally adapted with task-specific instructions to align them with the persuasion objective, while keeping their core methods unchanged.
Table~\ref{tab:baseline_prompt_adaptation} provides representative examples of the modifications applied to each baseline.
 
\begin{table*}[ht!]
\centering
\small
\begin{tabular}{p{1.8cm} p{6.5cm} p{6.5cm}}
\toprule
\textbf{Baseline} & \textbf{Original Instruction} & \textbf{Revised Instruction} \\
\midrule
\textsc{PAG}
& Write a summary, in English, of the research interests and topics of a researcher who has published the following papers.
& Write a summary of the most essential information about the author that would be useful for persuading or changing the author's view. \\
\midrule
\textsc{HSumm}
& System: Below is a transcript of a conversation between a human and an AI assistant that is intelligent and knowledgeable in psychology. \newline User: Hello! Please help me summarize the content of the conversation. \newline System: Sure, I will do my best to assist you.
& System: You are an expert assistant whose task is to extract information about the author from a set of passages. Your goal is to produce a compact, context-aware user profile optimized for changing the author's view. \\
\midrule
\textsc{RecurSumm}
& Remember, the memory should serve as a reference point to maintain continuity in the dialogue and help you respond accurately to the user based on their personality.
& Remember, the memory should serve as a reference point to produce a compact, context-aware user profile optimized for persuasive messaging toward the given post. \\
\bottomrule
\end{tabular}
\caption{Representative prompt modifications for each profiling baseline. Original instructions are adapted to the persuasion prediction task while preserving each method's core mechanism.}
\label{tab:baseline_prompt_adaptation}
\end{table*}

\section{Additional Experiment Results}
\label{app:additional_results}

\subsection{Comparison with Long-Context Direct Conditioning}
\label{sec:appendix_long_context}

To examine whether the proposed retrieval-and-profiling pipeline provides advantages beyond what long-context models can achieve through direct conditioning, we conduct additional experiments feeding raw user history directly to the predictor without any retrieval or profiling.
Specifically, the latest $k$ records are included for each user ($k = 10, 50, 100, 500$).

\begin{table}[t]
\centering
\small
\label{tab:long_context_baseline}
\begin{tabular}{llc}
\toprule
\textbf{Model} & \textbf{Setting} & \textbf{F1} \\
\midrule
\multirow{5}{*}{\texttt{Llama-3.3-70B}} 
  & $k=10$  & 0.3698 \\
  & $k=50$  & 0.2920 \\
  & $k=100$ & 0.2938 \\
  & $k=500$ & 0.2611 \\
  & \textbf{Ours} & \textbf{0.4661} \\
\midrule
\multirow{5}{*}{\texttt{Llama-3.1-8B}}
  & $k=10$  & 0.2411 \\
  & $k=50$  & 0.2257 \\
  & $k=100$ & 0.2629 \\
  & $k=500$ & 0.2621 \\
  & \textbf{Ours} & \textbf{0.4000} \\
\midrule
\multirow{5}{*}{\texttt{GPT-4o-mini}}
  & $k=10$  & 0.1344 \\
  & $k=50$  & 0.1126 \\
  & $k=100$ & 0.1383 \\
  & $k=500$ & 0.1038 \\
  & \textbf{Ours} & \textbf{0.2787} \\
\bottomrule
\end{tabular}
\caption{Prediction performance (F1) when directly conditioning on the latest $k$ user records, compared with our retrieval-and-profiling pipeline (\textbf{Ours}).}
\end{table}

Two observations emerge consistently across models.
First, directly feeding raw history performs substantially worse than our profiling-based approach across all settings, confirming that the benefit of our framework lies in the structured, persuasion-aware transformation of user history rather than simply providing more context.
Second, performance does not improve monotonically with the number of records and in some cases degrades as more records are added, suggesting that raw user history contains substantial noise and that indiscriminately feeding more records makes it harder for the predictor to identify persuasion-relevant signals.

We attribute this to a fundamental difference in what each approach does with user history.
Long-context direct conditioning effectively performs episodic memory retrieval---surfacing specific facts or events from a user's past---and leaves it to the predictor to extract useful signals from noisy, unstructured input.
In contrast, our framework goes beyond surface-level fact retrieval to \textbf{infer deeper user characteristics} such as values, beliefs, and reasoning styles, constructing a compact representation of the user's underlying disposition toward persuasion.
This distinction is particularly important for tasks involving human judgment and attitude change, where responses are driven not by isolated past events but by stable, abstract traits that must be inferred from behavioral traces.
The performance degradation observed with increasing history size further supports this view: more raw context does not help if the model cannot effectively abstract the persuasion-relevant signal from it.

Beyond prediction quality, directly conditioning on large user histories is also \textbf{highly inefficient} from an inference perspective, as the number of input tokens grows substantially with history size, resulting in significantly higher computational cost.
Together, these results demonstrate that long-context direct conditioning is neither an effective nor a practical alternative to our retrieval-and-profiling pipeline.

\subsection{Effect of Thread Context}
\label{sec:appendix_thread_context}
To examine the role of multi-turn discussion dynamics relative to persistent user characteristics, we conduct additional experiments on a test subset consisting of comments embedded in multi-turn exchanges. We compare three conditions under which the predictor determines whether the target comment receives a delta: (1) \textsc{User Profile} only, using profiles generated by our proposed framework; (2) \textsc{Thread Context} only, where we concatenate the turns prior to the target comment in the thread and provide them as additional context to the predictor alongside the target comment; and (3) \textsc{Profile + Thread Context}, combining both.

Table~\ref{tab:thread_context} reports comment-level view-change prediction F1 across the three predictor models. Two findings emerge consistently. First, user profile alone outperforms thread context alone across all predictors, indicating that modeling the persuadee's persistent characteristics is more informative than thread-level dynamics for this task. Second, combining user profile with thread context improves over thread context alone, suggesting that the two sources of information are complementary, though the profile remains the dominant signal.

\begin{table}[t]
\centering
\resizebox{\columnwidth}{!}{%
\begin{tabular}{lccc}
\toprule
\textbf{Setting} & \textbf{GPT} & \textbf{LLaMA-8B} & \textbf{LLaMA-70B} \\
\midrule
User Profile (Ours)      & 0.2054 & 0.3612 & 0.3755 \\
Thread Context           & 0.1437 & 0.2388 & 0.2680 \\
Profile + Thread Context & 0.2210 & 0.3186 & 0.3612 \\
\bottomrule
\end{tabular}%
}
\caption{Comment-level view-change prediction F1 under three conditioning settings: user profile only, thread context only, and their combination. GPT, LLaMA-8B, and LLaMA-70B denote \texttt{GPT-4o-mini}, \texttt{Llama-3.1-8B-Instruct}, and \texttt{Llama-3.3-70B-Instruct}, respectively.}
\label{tab:thread_context}
\end{table}

We note that our main experiments deliberately adopt a single-turn formulation restricted to top-level comments in order to isolate the contribution of user profiling as clearly as possible. By pairing each top-level comment with its corresponding original post, we disentangle the effect of the user's persistent characteristics from the contingent dynamics of a specific thread, providing a controlled setting in which the role of user profiling can be directly evaluated. Specifically, positive instances are top-level comments that directly received a delta, while negative instances are top-level comments from threads in which a delta was never awarded throughout the entire discussion. Extending the framework to incorporate multi-turn dynamics is a promising direction for future work.

\subsection{Effect of Stronger Predictor Models}
\label{sec:appendix_gpt5}
To examine whether the benefit of our framework persists under stronger predictor models, we conduct an additional experiment using GPT-5 as the predictor. Due to the prohibitive cost of retraining all pipeline components from scratch with a stronger model (approximately 10$\times$ the cost of GPT-4o-mini), we adopt a transfer setting in which the query generator and profiler trained under GPT-4o-mini are kept fixed, and only the predictor is replaced with GPT-5.

Table~\ref{tab:gpt5} reports F1 scores under three conditions: no personalization, retrieval-only, and our full framework. The results show a trend consistent with GPT-4o-mini: no personalization outperforms retrieval-only, yet applying our user profiles provides additional benefit over no personalization. This indicates that our profiler remains effective even under a stronger predictor.

\begin{table}[t]
\centering
\small
\begin{tabular}{lc}
\toprule
\textbf{Method} & \textbf{F1} \\
\midrule
No Personalization       & 0.3719 \\
Retrieval-only           & 0.1684 \\
\textbf{User Profile (Ours)} & \textbf{0.3791} \\
\bottomrule
\end{tabular}
\caption{View-change prediction F1 with GPT-5 as the predictor. The query generator and profiler are those trained under GPT-4o-mini and kept fixed; only the predictor is replaced.}
\label{tab:gpt5}
\end{table}

We note that this experiment is not a fully controlled comparison: the profiler and query generator were optimized under GPT-4o-mini's utility signal, and GPT-5 may prefer different records and profile dimensions. Since our framework derives training supervision directly from the target predictor's feedback, retraining the full pipeline with GPT-5 as the predictor would likely yield larger gains. We therefore view the current result as a conservative lower bound, and expect the performance gap to widen when all components are trained end-to-end with a stronger predictor.

\subsection{Ablation on Repetition Count}
\label{sec:appendix_ablation_repetition}

Because the predictor-side computation scales directly with the number of grouping repetitions $m$, we ablate $m$ to characterize the trade-off between additional compute and the stability of the training signal derived from the utility scores.

\paragraph{Compute scaling.}
As expected, the overall token usage increases near-linearly with $m$:
$m=5$ requires approximately 2.3B tokens (2{,}338{,}263{,}616), corresponding to $1.67\times$ the compute of $m=3$;
$m=10$ requires approximately 4.6B tokens (4{,}682{,}146{,}776), corresponding to $3.34\times$ the compute of $m=3$.

\paragraph{Compute-stability trade-off.}
The query generator is trained with DPO using preference pairs of candidate queries, where the ``chosen'' and ``rejected'' labels are derived from their NDCG@5 scores (computed by using record utilities as graded relevance).
For our analysis, we bucket candidate queries into chosen vs.\ rejected pools using the same thresholds used in our experiments (chosen if NDCG@5 $\geq 0.55$, rejected if $\leq 0.45$), and measure stability via a \textbf{flip rate}: among all candidates that are labeled chosen or rejected under $m=3$, the fraction of candidates whose label changes when utility scores are recomputed with larger $m$.
The flip rates are $5.91\%$ for $m=3 \to 5$ and $8.05\%$ for $m=3 \to 10$.

Considering that two candidates constitute a preference sample, this means that only $4.025\%$ of preference samples (at most) have their original preference direction reversed as repetitions increase from 3 to 10, while $\geq 92\%$ of preference/rejection labels used for DPO are preserved.
Given that moving from $m=3$ to $m=10$ costs $>3\times$ more compute, these results indicate that $m=3$ is a compute-efficient choice that retains the vast majority of the training signal.

\subsection{Additional Retrieval Results}
\label{sec:appendix_retrieval_additional}

Table~\ref{tab:appendix_retrieval} reports retrieval performance under different predictor-specific
persuasion utility signals, using the same candidate record pool and query strategies as in the main experiment
(Table~\ref{tab:retrieval_main}).

\begin{table}[t]
\centering
\small
\setlength{\tabcolsep}{5pt}
\resizebox{\linewidth}{!}{
\begin{tabular}{lcccc}
\toprule
& \multicolumn{2}{c}{\textbf{Llama-3.3-70B-Instruct}} 
& \multicolumn{2}{c}{\textbf{GPT-4o-mini}} \\
\cmidrule(lr){2-3} \cmidrule(lr){4-5}
\textbf{Query Strategy}
& \textbf{NCG@5} & \textbf{NDCG@5}
& \textbf{NCG@5} & \textbf{NDCG@5} \\
\midrule
Random   & 0.7461 & 0.7395 & 0.4713 & 0.4647 \\
BGE-Post & 0.7461 & 0.7357 & 0.4826 & 0.4736 \\
HyDE     & 0.7528 & \textbf{0.7482} & 0.4685 & 0.4562 \\
\midrule
\textit{Ours}
& \textbf{0.7536} & 0.7471
& \textbf{0.4827} & \textbf{0.4747} \\
\bottomrule
\end{tabular}
}
\caption{Additional retrieval-side results under different predictor-specific persuasion utility signals.
All methods are evaluated on the same record pool as the main experiment.
Random reports the average over 10 runs.}
\label{tab:appendix_retrieval}
\end{table}

\begin{table*}[t]
\centering
\small
\setlength{\tabcolsep}{6pt}
\begin{tabular}{lccc ccc ccc}
\toprule
& \multicolumn{3}{c}{Llama-3.1-8B-Instruct}
& \multicolumn{3}{c}{Llama-3.3-70B-Instruct}
& \multicolumn{3}{c}{GPT-4o-mini} \\
\cmidrule(lr){2-4} \cmidrule(lr){5-7} \cmidrule(lr){8-10}
\textbf{Retrieval} 
& Demograph. & Base & \textit{Ours}
& Demograph. & Base & \textit{Ours}
& Demograph. & Base & \textit{Ours} \\
\midrule
Recent  
& 0.5828 & 0.5953 & 0.6088
& 0.6577 & 0.6740 & 0.6746
& 0.6214 & 0.6214 & 0.6232 \\

Random  
& \textbf{0.5859} & \textbf{0.6121} & 0.6112
& 0.6528 & 0.6669 & 0.6716
& 0.6188 & 0.6121 & 0.6253 \\

BM25    
& 0.5858 & 0.5955 & 0.6082
& 0.6564 & 0.6588 & 0.6697
& 0.6159 & 0.6189 & 0.6365 \\

BGE  
& 0.5851 & 0.5859 & 0.6029
& \textbf{0.6596} & \textbf{0.6768} & 0.6798
& 0.6226 & 0.6305 & 0.6349 \\

HyDE    
& 0.5845 & 0.5997 & 0.6104
& 0.6569 & 0.6655 & 0.6825
& 0.6216 & \textbf{0.6311} & \textbf{0.6447} \\

\midrule
\textit{Ours}
& 0.5850 & 0.6054 & \textbf{\underline{0.6146}}
& 0.6574 & 0.6736 & \textbf{\underline{0.6828}}
& \textbf{0.6232} & 0.6020 & \underline{0.6299} \\
\bottomrule
\end{tabular}
\caption{Effect of retriever and profiler choices on view-change prediction under different predictors (AUC).
Random reports the average performance over 10 runs.
\underline{Underlined} results denote our final proposed method, while \textbf{boldface} highlights the best-performing configuration within each column.
Column groups correspond to different predictor models, with sub-columns indicating profiler configurations (demographic, base profiler, and our trained profiler).
}
\label{tab:e2e_results_auc}
\end{table*}

\subsection{AUC scores across different retrieval and profiling variants}
\label{appendix_AUC}

Table~\ref{tab:e2e_results_auc} reports the full end-to-end AUC results across different retriever and profiler combinations for each predictor model.

\section{Details of Profiler Analysis}
\label{sec:profiler_analysis_details}

\subsection{Additional Results}

\begin{figure}[t]
    \centering
    \includegraphics[width=1.0\linewidth]{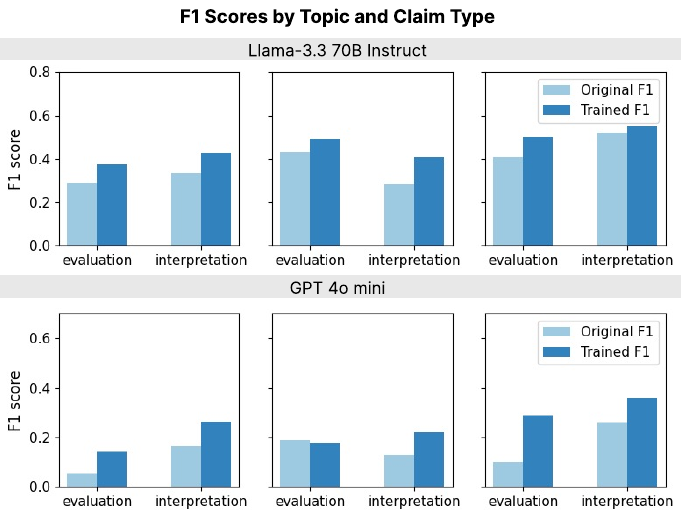}
    \caption{F1 by topic and claim type of the post, comparing the original and trained profilers.
}
    \label{fig:profiler-analysis-1-full}
\end{figure}

Figure~\ref{fig:profiler-analysis-1-full} extends the main-text analysis by reporting F1 scores by topic and claim type for \texttt{GPT-4o-mini} and \texttt{Llama-3.3-70B-Instruct}, comparing the original and trained profilers.

\begin{figure}[t]
    \centering
    \includegraphics[width=\linewidth]{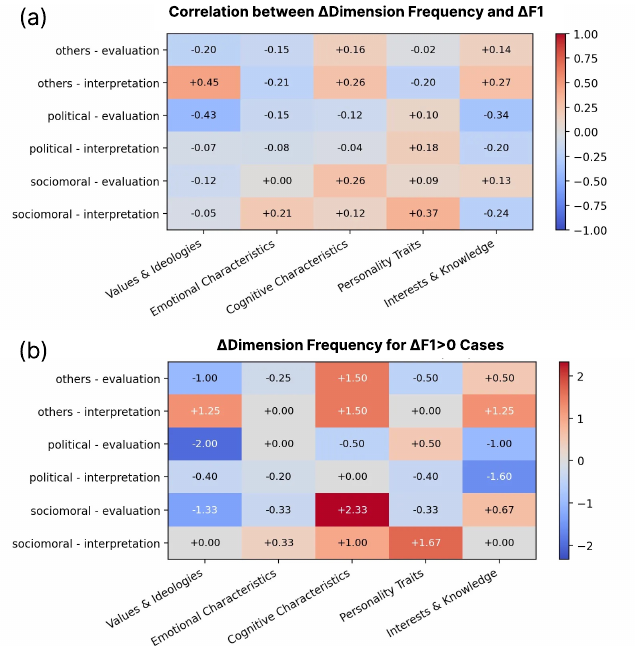}
    \caption{
Analysis of profile-dimension frequency shifts ($\Delta$DF) and performance gains ($\Delta$F1) between the original and trained profilers.
(a) Correlation between $\Delta$DF and $\Delta$F1.
(b) $\Delta$DF for cases with $\Delta$F1 $>$ 0.
\texttt{GPT-4o-mini} is used as the predictor.
}
    \label{fig:profiler-analysis-2-gpt}
\end{figure}

\begin{figure}[t]
    \centering
    \includegraphics[width=\linewidth]{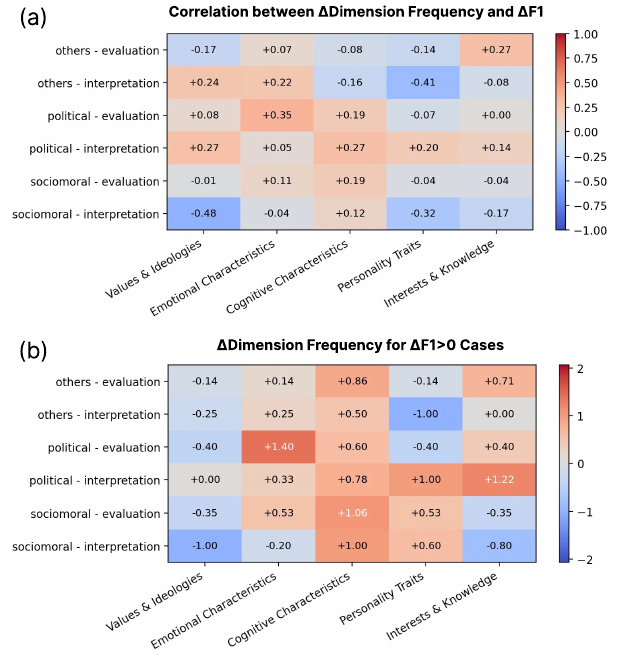}
    \caption{
Analysis of profile-dimension frequency shifts ($\Delta$DF) and performance gains ($\Delta$F1) between the original and trained profilers.
(a) Correlation between $\Delta$DF and $\Delta$F1.
(b) $\Delta$DF for cases with $\Delta$F1 $>$ 0.
\texttt{Llama-3.3-70B-Instruct} is used as the predictor.
}
    \label{fig:profiler-analysis-2-70b}
\end{figure}

Figures~\ref{fig:profiler-analysis-2-gpt} and~\ref{fig:profiler-analysis-2-70b} present the profile-dimension analysis results for \texttt{GPT-4o-mini} and \texttt{Llama-3.3-70B-Instruct}, respectively.

\subsection{Failure Case Analysis}
\label{sec:failure_case_analysis}
To better understand failure cases, we conduct an additional analysis analogous to Figure~\ref{fig:profiler-analysis-2}b, focusing on instances where $\Delta\text{F1} < 0$ (i.e., the trained profile underperforms the original).
Using the same procedure as in the profiler analysis (Section~\ref{sec:profiler_analysis}), we compute delta-dimension frequency for performance-degrading profiles, using \texttt{Llama-3.1-8B-Instruct} as the predictor.
Table~\ref{tab:delta_dim_freq_decrease} reports the delta-dimension frequency for these cases.
The results show that the shifts are post-type dependent: \textbf{no single dimension is uniformly over- or under-produced across all post types}, indicating that failures are not driven by a global bias toward any particular dimension.
To interpret these shifts, Table~\ref{tab:pearson_corr_delta} reproduces the Pearson correlation between delta-dimension frequency and performance change.
As is largely expected, the results indicate that failures arise from misweighting profile dimensions: in contrast to performance-improving profiles, they do not sufficiently increase dimensions positively correlated with gains for a given post type and instead overrepresent dimensions negatively correlated with performance.
A concrete example appears in Others--Interpretation posts.
In this category, \textit{Interest \& Knowledge} ($+0.39$) and \textit{Personality Traits} ($+0.52$) are strongly positively correlated with performance gains (Table~\ref{tab:pearson_corr_delta}).
However, performance-degrading profiles contain on average approximately two fewer \textit{Interest \& Knowledge} items and one fewer \textit{Personality Traits} item than the original profiles (Table~\ref{tab:delta_dim_freq_decrease}), indicating that under-representation of beneficial dimensions contributes to the performance drop.

\begin{table}[ht!]
\centering
\small
\resizebox{\columnwidth}{!}{%
\begin{tabular}{lccccc}
\toprule
\textbf{Task} & \textbf{Val} & \textbf{Emo} & \textbf{Cog} & \textbf{Pers} & \textbf{Int} \\
\midrule
Others-Eval       & +0.14 & +0.00 & +0.29 & +0.71 & $-$0.57 \\
Others-Interp     & +1.00 & +0.00 & +0.00 & $-$1.00 & $-$2.00 \\
Political-Eval    & +0.00 & +0.33 & +1.00 & +0.67 & +0.67 \\
Political-Interp  & +0.14 & +0.57 & +0.29 & $-$0.29 & +1.00 \\
Sociomoral-Eval   & +0.50 & +0.40 & $-$0.50 & $-$0.70 & +0.70 \\
Sociomoral-Interp & +0.33 & $-$1.00 & +1.00 & $-$0.67 & +0.67 \\
\bottomrule
\end{tabular}%
}
\caption{Delta-dimension frequency ($\Delta$DF) for performance-decreasing cases ($\Delta$F1 $< 0$). Column abbreviations: Val = Values \& Ideologies; Emo = Emotional Characteristics; Cog = Cognitive Characteristics; Pers = Personality Traits; Int = Interests \& Knowledge.}
\label{tab:delta_dim_freq_decrease}
\end{table}

\begin{table}[ht!]
\centering
\small
\resizebox{\columnwidth}{!}{%
\begin{tabular}{lccccc}
\toprule
\textbf{Task} & \textbf{Val} & \textbf{Emo} & \textbf{Cog} & \textbf{Pers} & \textbf{Int} \\
\midrule
Others-Eval       & +0.28 & +0.11 & +0.05 & $-$0.32 & +0.27 \\
Others-Interp     & $-$0.17 & $-$0.26 & $-$0.01 & +0.52 & +0.39 \\
Political-Eval    & $-$0.14 & $-$0.20 & $-$0.29 & +0.13 & $-$0.18 \\
Political-Interp  & $-$0.13 & $-$0.32 & +0.02 & +0.21 & $-$0.04 \\
Sociomoral-Eval   & $-$0.12 & +0.00 & +0.22 & +0.10 & $-$0.10 \\
Sociomoral-Interp & $-$0.55 & +0.13 & +0.17 & +0.33 & $-$0.29 \\
\bottomrule
\end{tabular}%
}
\caption{Pearson correlation between $\Delta$DF and $\Delta$F1. Abbreviations follow Table~\ref{tab:delta_dim_freq_decrease}.}
\label{tab:pearson_corr_delta}
\end{table}

\section{Details of User Record Analysis}
\label{sec:record-analysis}

\subsection{Additional Results}

\begin{figure}[t]
    \centering
    \includegraphics[width=\linewidth]{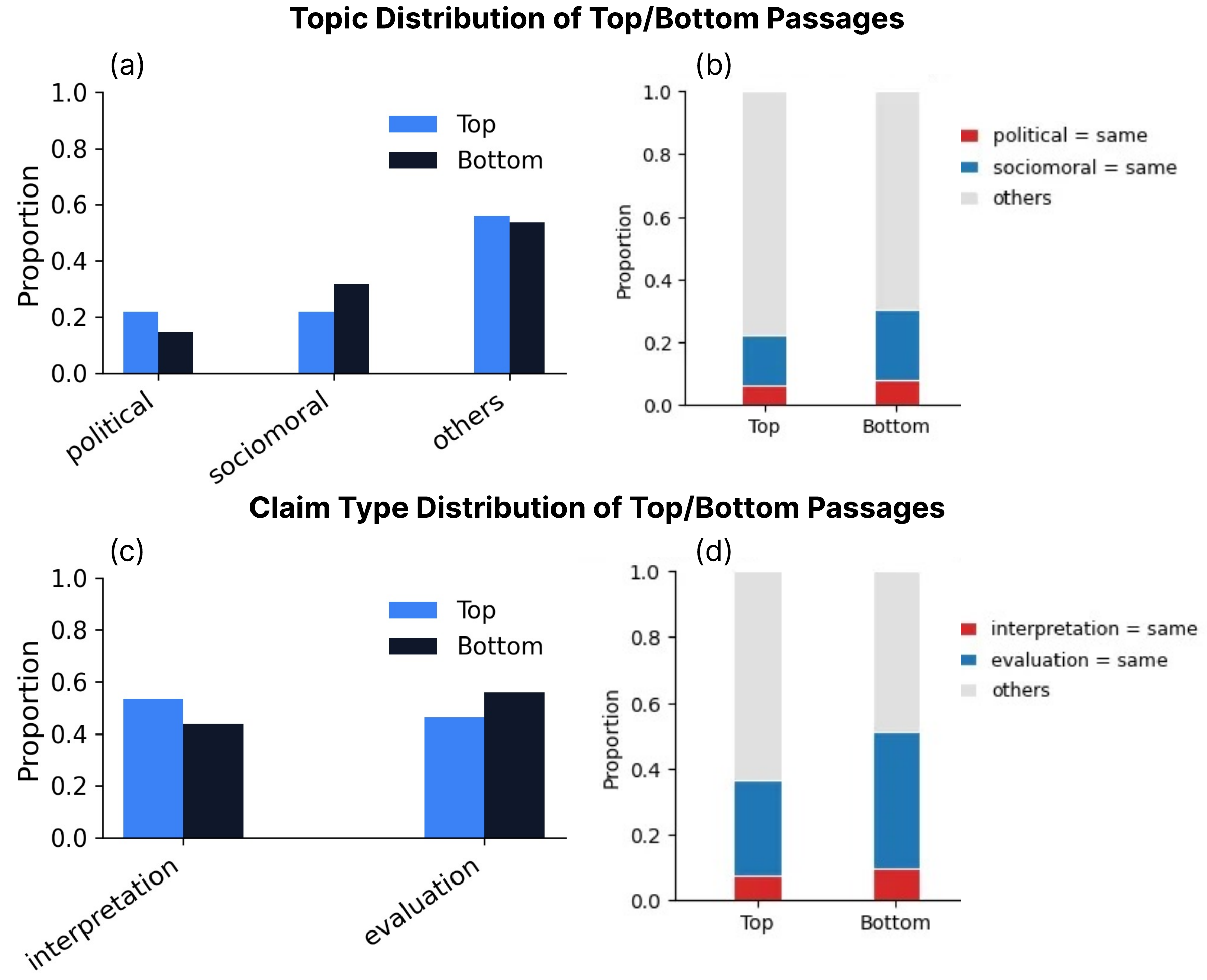}
    \caption{Topic and claim type distributions of top- and bottom-ranked records.
(a,c) Marginal distributions by topic and claim type.
(b,d) Proportions of records sharing the same topic or claim type with the post.}
    \label{fig:record-analysis-1}
\end{figure}

Figure~\ref{fig:record-analysis-1} analyzes the topical and claim type characteristics of user records ranked by utility score.
Top-ranked records tend to align more closely with the post in both topic and claim type, whereas bottom-ranked records show weaker alignment.

Table~\ref{tab:score_distribution_mean} reports mean F1 over all records as well as the top-5 and bottom-5 subsets, showing that while \texttt{Llama-3.3-70B-Instruct} assigns higher scores overall, \texttt{GPT-4o-mini} exhibits a larger contrast between high- and low-ranked records.
This trend is further reflected in Table~\ref{tab:margin_distribution}, where \texttt{GPT-4o-mini} yields consistently larger margins between top-5 and bottom-5 records.

\begin{table}[t]
\centering
\small
\begin{tabular}{lccc}
\toprule
\textbf{Model} 
& \textbf{All} 
& \textbf{Top-5} 
& \textbf{Bottom-5} \\
\midrule
GPT-4o-mini 
& 0.163 
& 0.302 
& 0.092 \\

Llama-3.3-70B-Instruct 
& \textbf{0.375} 
& \textbf{0.468} 
& \textbf{0.301} \\

Llama-3.1-8B-Instruct 
& 0.248 
& 0.321 
& 0.183 \\
\bottomrule
\end{tabular}
\caption{Mean F1 scores across all records, top-5 records,
and bottom-5 records.
While \texttt{Llama-3.3-70B-Instruct} assigns higher absolute scores overall,
\texttt{GPT-4o-mini} exhibits a larger contrast between top-5 and bottom-5 records.}
\label{tab:score_distribution_mean}
\end{table}

\begin{table}[t]
\centering
\small
\begin{tabular}{lccc}
\toprule
\textbf{Model} 
& \textbf{Mean} 
& \textbf{Median (p50)} 
& \textbf{p90} \\
\midrule
GPT-4o-mini 
& \textbf{0.180} 
& \textbf{0.133} 
& \textbf{0.500} \\

Llama3.3-70B-Instruct
& 0.135 
& 0.090 
& 0.337 \\

Llama-3.1-8B-Instruct 
& 0.112 
& 0.080 
& 0.260 \\
\bottomrule
\end{tabular}
\caption{Distribution of margin between high- and low-scoring records (min top-5 minus max bottom-5).
\texttt{GPT-4o-mini} exhibits consistently larger margins, indicating clearer separation between beneficial and non-beneficial records.}
\label{tab:margin_distribution}
\end{table}

\section{Details of Efficiency Analysis}
\label{sec:appendix_efficiency}

This appendix provides per-stage breakdowns and measurement details
for the efficiency analysis reported in
Section~\ref{sec:efficiency}.

\subsection{Measurement Setup}

We measure per-instance inference cost under a realistic deployment
scenario in which user history embeddings are maintained offline,
so that only newly added records trigger re-encoding at inference
time. Tokens are counted directly from the actual prompts and
completions consumed by each LLM call in the pipeline. FLOPs are
estimated from token counts using the standard approximation
$C \approx 2 N T$, where $N$ is the number of model parameters and
$T$ is the number of tokens processed. All measurements are
averaged over the test set with \texttt{Llama-3.1-8B-Instruct} as
the predictor, profiler, and query generator.

\subsection{Per-Stage Token Usage and FLOPs}

Tables~\ref{tab:appendix_tokens} and~\ref{tab:appendix_flops}
decompose per-instance cost into three stages: search (query
formulation and retrieval call), profiling (profile construction
from retrieved or full history), and prediction (final view-change
prediction). Search tokens correspond to the query-generation step
in our method and to BM25 query formulation in \textsc{PAG}.
\textsc{HSumm} and \textsc{RecurSumm} operate directly on full user
history without a separate retrieval query, so their search cost is
zero. Profiling dominates the cost of full-history baselines, while
retrieval-based methods distribute cost more evenly across query
generation, profiling, and prediction.

\begin{table}[t]
\centering
\small
\setlength{\tabcolsep}{3pt}
\begin{tabular}{lrrrr}
\toprule
\textbf{Method} & \textbf{Search} & \textbf{Profile} & \textbf{Predict} & \textbf{Total} \\
\midrule
\textsc{PAG}       & 294    & 1{,}464  & 1{,}470 & 3{,}228 \\
\textbf{Ours}      & 916    & 1{,}479  & 1{,}034 & 3{,}429 \\
\textsc{RecurSumm} & 0      & 20{,}750 & 928     & 21{,}678 \\
\textsc{HSumm}     & 0      & 41{,}816 & 1{,}064 & 42{,}880 \\
\bottomrule
\end{tabular}
\caption{Per-instance token usage across stages.}
\label{tab:appendix_tokens}
\end{table}

\begin{table}[t]
\centering
\small
\setlength{\tabcolsep}{3pt}
\resizebox{\columnwidth}{!}{%
\begin{tabular}{lrrrr}
\toprule
\textbf{Method} & \textbf{Search} & \textbf{Profile} & \textbf{Predict} & \textbf{Total} \\
\midrule
\textsc{PAG}       & $1.2{\times}10^{6}$  & $2.30{\times}10^{13}$ & $2.41{\times}10^{13}$ & $4.71{\times}10^{13}$ \\
\textbf{Ours}      & $9.97{\times}10^{12}$ & $2.41{\times}10^{13}$ & $1.68{\times}10^{13}$ & $5.09{\times}10^{13}$ \\
\textsc{RecurSumm} & 0                    & $3.43{\times}10^{14}$ & $1.50{\times}10^{13}$ & $3.58{\times}10^{14}$ \\
\textsc{HSumm}     & 0                    & $6.86{\times}10^{14}$ & $1.73{\times}10^{13}$ & $7.03{\times}10^{14}$ \\
\bottomrule
\end{tabular}%
}
\caption{Per-instance FLOPs across stages.}
\label{tab:appendix_flops}
\end{table}

\subsection{Training Cost of Persuasion Utility Scoring}

Record-level persuasion utility scoring
(Section~\ref{sec:persuasion_utility_scoring}) is performed once during
training and requires repeated profile generation and prediction
over randomly grouped user records. Under our default setting of
three repetitions with \texttt{GPT-4o-mini} as the predictor, the
total token consumption on the training set is approximately $362$M
profiler tokens and $1.04$B predictor tokens, totaling $1.40$B
tokens. Predictor-side computation dominates ($74.2\%$ of total),
reflecting that each sampled profile is evaluated against multiple
comments for the associated post. This cost is incurred only once
during training; at inference time, no utility scoring is needed,
and the framework applies to new users without additional training.

\end{document}